%% file: main.tex
\documentclass[10pt,journal,compsoc]{IEEEtran}
\ifCLASSOPTIONcompsoc
  \usepackage[nocompress]{cite}
\else
  \usepackage{cite}
\fi

\usepackage{amsmath,amssymb,amsfonts}
\usepackage{bm}
\usepackage{dsfont}
\usepackage{graphicx}
\usepackage{adjustbox}
\usepackage[bookmarks=false]{hyperref}
\usepackage{comment}
\usepackage{algorithm}
\usepackage{algpseudocode}
\usepackage{units}
\usepackage{multirow}
\usepackage{array}
\usepackage{booktabs}

\usepackage{color}
\definecolor{darkspringgreen}{rgb}{0.09, 0.45, 0.27}

\usepackage[normalem]{ulem}
\newcommand{\rev}[1]{#1}
\newcommand{\revRemove}[1]{}
\newcommand{\revFloatRemove}[1]{}
\newcommand{\revEqRemove}[1]{}

\usepackage{tikz}
\usepackage{textcomp}
\usepackage[doipre={DOI:~}]{uri}
\usepackage{lipsum}

\newcommand\copyrighttext{%
  \footnotesize \textcopyright 2022 IEEE. Personal use of this material is permitted.  Permission from IEEE must be obtained for all other uses, in any current or future media, including reprinting/republishing this material for advertising or promotional purposes, creating new collective works, for resale or redistribution to servers or lists, or reuse of any copyrighted component of this work in other works.
 
  Accepted at IEEE Transactions on Computers.
  ~\doi{10.1109/TC.2022.3177955} }
\newcommand{\copyrightnotice}{%
\begin{tikzpicture}[remember picture,overlay,scale=0.80, every node/.style={scale=0.80}]
\node[anchor=south,yshift=10pt] at (current page.south) {\fbox{\parbox{\dimexpr\textwidth-\fboxsep-\fboxrule\relax}{\copyrighttext}}};
\end{tikzpicture}%
}

\begin{document}
\bstctlcite{IEEEexample:BSTcontrol}
\title{Lightweight Neural Architecture Search for Temporal Convolutional Networks at the Edge}
\author{Matteo~Risso,
        Alessio~Burrello,~\IEEEmembership{Member,~IEEE,}
        Francesco~Conti,~\IEEEmembership{Member,~IEEE,}
        Lorenzo~Lamberti,~\IEEEmembership{Member,~IEEE,}
        Yukai~Chen,~\IEEEmembership{Member,~IEEE,}
        Luca~Benini,~\IEEEmembership{Fellow,~IEEE,}
        Enrico~Macii,~\IEEEmembership{Fellow,~IEEE,}
        Massimo~Poncino,~\IEEEmembership{Fellow,~IEEE,}
        and~Daniele~Jahier~Pagliari,~\IEEEmembership{Member,~IEEE}% <-this % stops a space
\IEEEcompsocitemizethanks{\IEEEcompsocthanksitem M. Risso, Y. Chen, M. Poncino and D. Jahier Pagliari are with the Department
of Control and Computer Engineering, Politecnico di Torino, 10129, Turin, Italy. E-mail: name.firstsurname@polito.it
\IEEEcompsocthanksitem E. Macii is with the Interuniversity Department
of Regional and Urban Studies and Planning, Politecnico di Torino, 10129, Turin, Italy. E-mail: enrico.macii@polito.it
\IEEEcompsocthanksitem A. Burrello, F. Conti, L. Lamberti and L. Benini are with the Department of Electrical, Electronic, and Information Engineering, University of Bologna, 40136 Bologna, Italy. E-mail: name.surname@unibo.it
\IEEEcompsocthanksitem L. Benini is also with the  Department  of  Information  Technology  and Electrical Engineering at the ETH Zurich, 8092 Zurich, Switzerland. E-mail:lbenini@iis.ee.ethz.ch
}% <-this % stops a space
\thanks{Manuscript received January XX, XXXX; revised January XX, XXXX.}}

\markboth{IEEE Transactions on Computers,~Vol.~XX, No.~X, January~XXXX}%
{Risso \MakeLowercase{\textit{et al.}}: Lightweight Neural Architecture Search for Temporal Convolutional Networks at the Edge}

\IEEEtitleabstractindextext{%
\begin{abstract}
\input{sections/00_Abstract}
\end{abstract}

}

\maketitle

\copyrightnotice

\IEEEdisplaynontitleabstractindextext
\IEEEpeerreviewmaketitle

\ifCLASSOPTIONcompsoc
\IEEEraisesectionheading{\section{Introduction}\label{sec:introduction}}
\else
\section{Introduction}
\label{sec:introduction}
\fi
\input{sections/01_Introduction}

\section{Background and Related Works} \label{sec:back_related}
\input{sections/02_Background}
\section{Proposed Method} \label{sec:methods}
\input{sections/03_Method}

\section{Benchmarks} \label{sec:benchmark}
\input{sections/04_Benchmark}

\section{Experimental Results} \label{sec:results}
\input{sections/05_Results}

\section{Conclusion}
\input{sections/06_Conclusions} \label{sec:conclusion}

% \appendices

% % you can choose not to have a title for an appendix
% % if you want by leaving the argument blank
% \section{}

% \ifCLASSOPTIONcompsoc
%   % The Computer Society usually uses the plural form
%   \section*{Acknowledgments}
% \else
%   % regular IEEE prefers the singular form
% \fi

% The authors would like to thank...\rev{TODO!!}

\ifCLASSOPTIONcaptionsoff
  \newpage
\fi

\bibliographystyle{IEEEtran}
\bibliography{bstctl,references}

\begin{IEEEbiography}[{\includegraphics[width=1in,height=1.25in,clip,keepaspectratio]{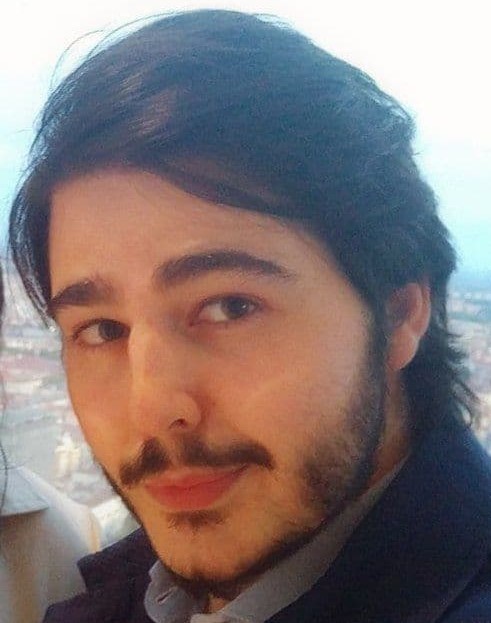}}]{Matteo Risso}
received his B.Sc degree in Physical Engineering and M.Sc degree in Electronic Engineering at the Politecnico di Torino, Italy, in 2018 and 2020. He is currently working toward his Ph.D. degree at Politecnico di Torino, Italy. His research interests include Embedded Machine Learning and Energy-Efficient Embedded Systems.
\end{IEEEbiography}

\begin{IEEEbiography}[{\includegraphics[width=1in,height=1.25in,clip,keepaspectratio]{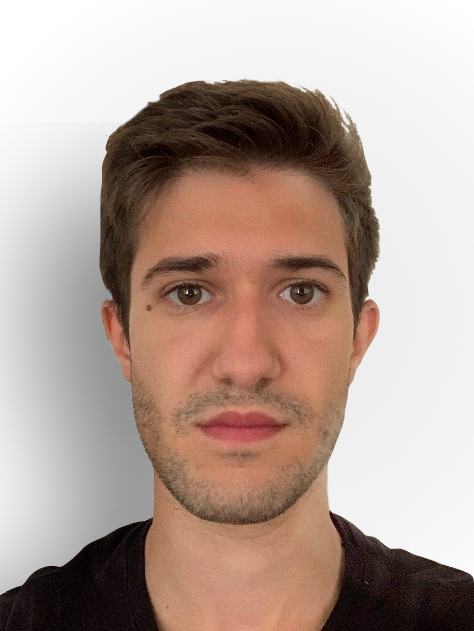}}]{Alessio Burrello}
received his B.Sc and M.Sc degree in Electronic Engineering at the Politecnico of Turin, Italy, in 2016 and 2018.  He is currently working toward his Ph.D. degree at the Department of Electrical, Electronic and Information Technologies Engineering (DEI) of the University of Bologna, Italy.
His research interests include parallel programming models for embedded systems, machine and deep learning, hardware oriented deep learning, and code optimization for multi-core systems.
\end{IEEEbiography}

\begin{IEEEbiography}[{\includegraphics[width=1in,height=1.25in,clip,keepaspectratio]{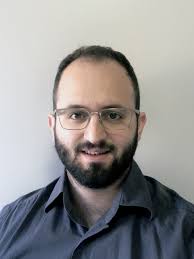}}]{Francesco Conti} received the Ph.D. degree in electronic engineering from the University of Bologna, Italy, in 2016. He is currently an Assistant Professor in the DEI Department of the University of Bologna. 
From 2016 to 2020, he held a research grant in the DEI department of University of Bologna and a position as postdoctoral researcher at the Integrated Systems Laboratory of ETH Zurich in the Digital Systems group.
His research focuses on the development of deep learning based intelligence on top of ultra-low power, ultra-energy efficient programmable Systems-on-Chip.
His research work has resulted in more than 60 publications in international conferences and journals and has been awarded several times, including the 2020 IEEE TCAS-I Darlington Best Paper Award.
\end{IEEEbiography}

\begin{IEEEbiography}[{\includegraphics[width=1in,height=1.25in,clip,keepaspectratio]{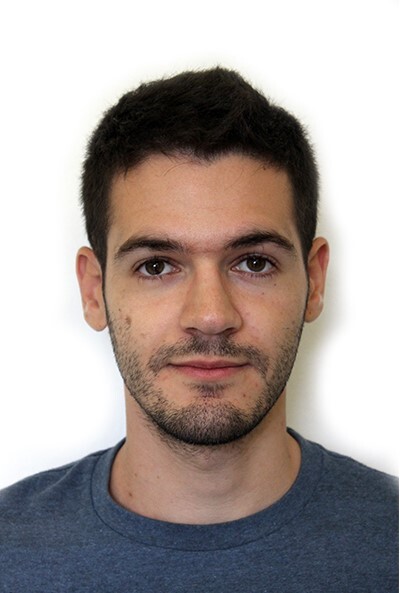}}]{Lorenzo Lamberti} graduated with honors in both Bachelor's (2016) and Master's (2019) degree at the University of Bologna, Italy, where he is now pursuing a Ph.D. in Electronic Engineering in the Energy-Efficient Embedded Systems Laboratory. Previously, he has been an intern at the Fermi National Accelerator Laboratory of Chicago, US, and at the Datalogic Artificial Intelligence Laboratory in Pasadena, US. His research is currently targeted at autonomous navigation for nano-size unmanned aerial vehicles. In particular, he focuses on neural architecture search, training and in-field deployment of Deep Neural Network-based navigation tasks on low-power MCUs.
\end{IEEEbiography}

\begin{IEEEbiography}[{\includegraphics[width=1in,height=1.25in,clip,keepaspectratio]{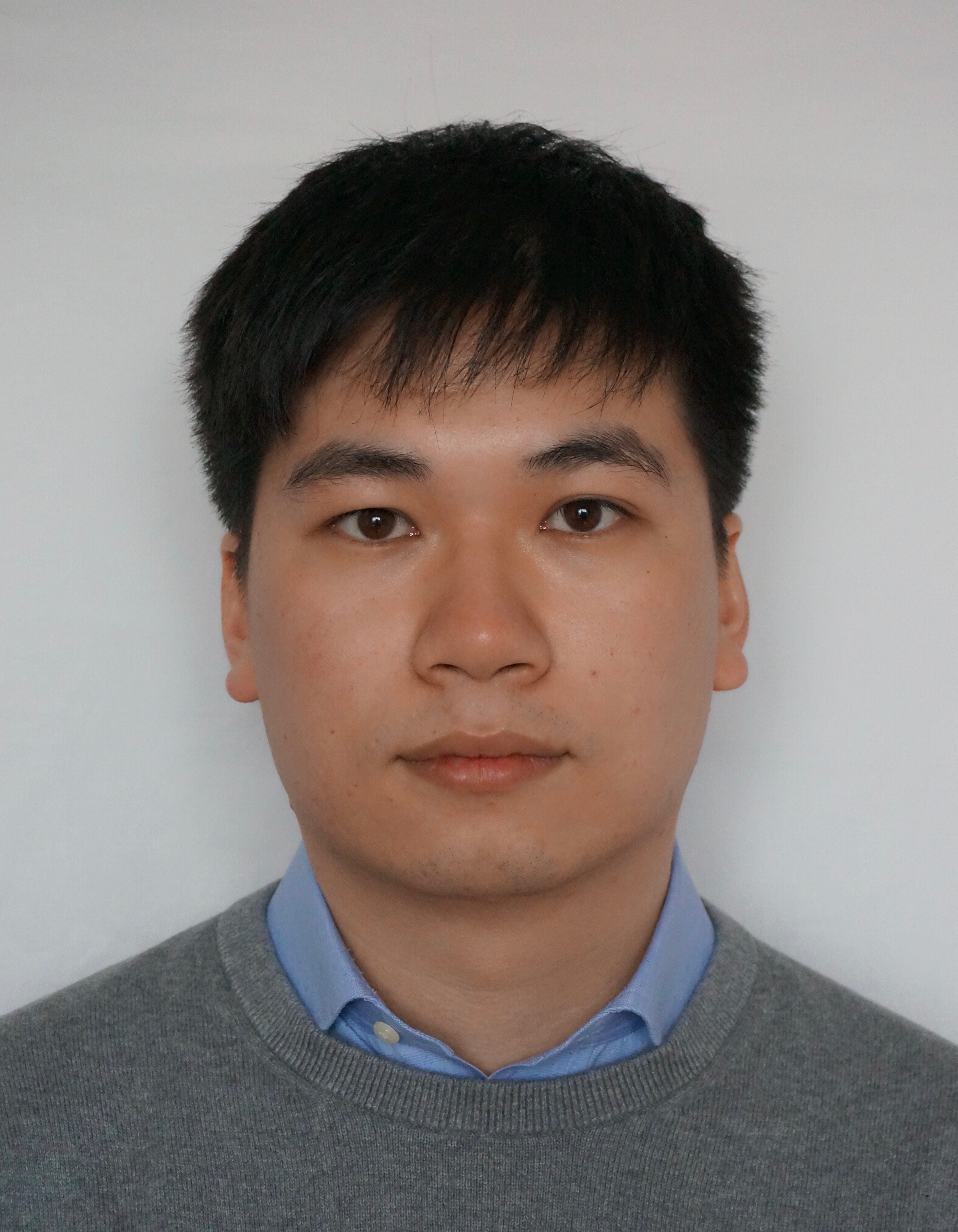}}]
{Yukai Chen} received the M.Sc. and Ph.D. degrees in computer engineering from Politecnico di Torino, Turin, Italy, in 2014 and 2018, respectively. He is currently a Postdoc researcher at Politecnico di Torino. His research interest includes computer-aided design for non-functional properties (power, thermal, reliability) modelling, simulation, and optimization of Cyber-Physical Systems, emphasizing low-power design, design automation and design space exploration.
\end{IEEEbiography}

\begin{IEEEbiography}[{\includegraphics[width=1in,height=1.25in,clip,keepaspectratio]{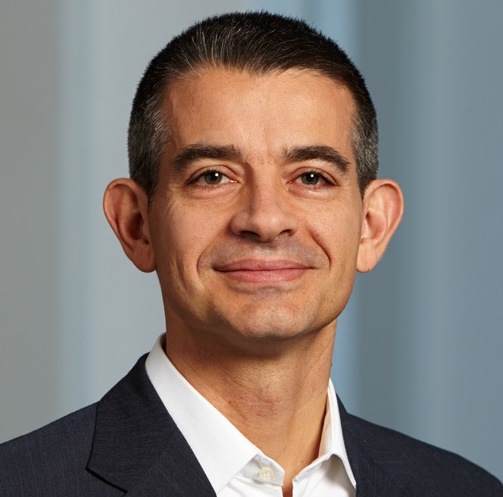}}]{Luca Benini}
is the Chair of Digital Circuits and Systems at ETH Z\"urich and a Full Professor at the University of Bologna.
He has served as Chief Architect for the Platform2012 in STMicroelectronics, Grenoble.
Dr. Benini’s research interests are in energy-efficient systems and multi-core SoC design. 
He is also active in the area of energy-efficient smart sensors and sensor networks. 
He has published more than 1’000 papers in peer-reviewed international journals and conferences, five books and several book chapters. 
He is a Fellow of the ACM and a member of the Academia Europaea.
\end{IEEEbiography}

\begin{IEEEbiography}[{\includegraphics[width=1in,height=1.25in,clip,keepaspectratio]{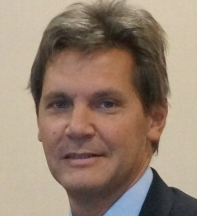}}]{Enrico Macii} is a Full Professor of Computer Engineering with the Politecnico di Torino, Torino, Italy. He holds a Laurea degree in electrical engineering from the Politecnico di Torino, a Laurea degree in computer science from the Università di Torino, Turin, and a PhD degree in computer engineering from the Politecnico di Torino. His research interests are in the design of digital electronic circuits and systems, with a particular emphasis on low-power consumption aspects energy efficiency, sustainable urban mobility, clean and intelligent manufacturing. He is a Fellow of the IEEE.
\end{IEEEbiography}

\begin{IEEEbiography}[{\includegraphics[width=1in,height=1.25in,clip,keepaspectratio]{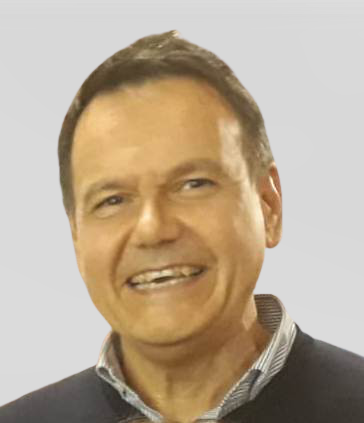}}]{Massimo Poncino} is a Full Professor of Computer Engineering with the Politecnico di Torino, Torino, Italy. His current research interests include various aspects of design automation of digital systems, with emphasis on the modeling and optimization of energy-efficient systems. He received a PhD in computer engineering and a Dr.Eng. in electrical engineering from Politecnico di Torino. He is a Fellow of the IEEE.
\end{IEEEbiography}

\begin{IEEEbiography}[{\includegraphics[width=1in,height=1.25in,clip,keepaspectratio]{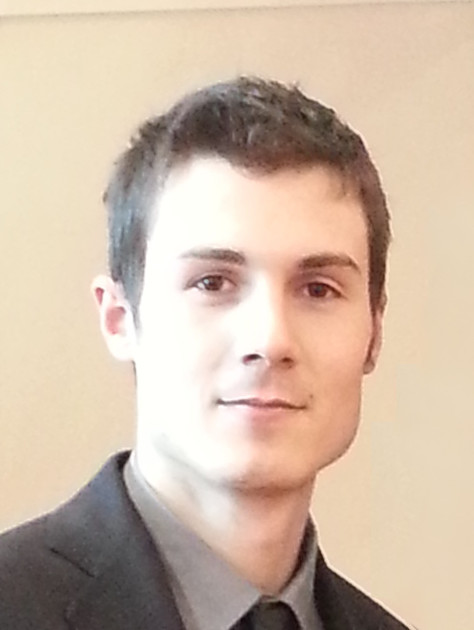}}]{Daniele Jahier Pagliari} received the M.Sc. and Ph.D. degrees in computer engineering from the Politecnico di Torino, Turin, Italy, in 2014 and 2018, respectively. He is currently an Assistant Professor with the Politecnico di Torino. His research interests are in the computer-aided design and optimization of digital circuits and systems, with a particular focus on energy-efficiency aspects and on emerging applications, such as machine learning at the edge.
\end{IEEEbiography}

\end{document}

%% file: sections/00_Abstract.tex
Neural Architecture Search (NAS) is quickly becoming the go-to approach to optimize the structure of Deep Learning (DL) models for complex tasks such as Image Classification or Object Detection.
However, many other relevant applications of DL, especially at the edge, are based on time-series processing and require models with unique features, for which NAS is less explored.
This work focuses in particular on Temporal Convolutional Networks (TCNs), a convolutional model for time-series processing that has recently emerged as a promising alternative to more complex recurrent architectures. 
We propose the first NAS tool that explicitly targets the optimization of the most peculiar architectural parameters of TCNs, namely dilation, receptive-field and number of features in each layer. 
The proposed approach searches for networks that offer good trade-offs between accuracy and number of parameters/operations, enabling an efficient deployment on embedded platforms.
Moreover, its fundamental feature is that of being \textit{lightweight} in terms of search complexity, making it usable even with limited hardware resources.
\revRemove{With respect to ProxylessNAS, a state-of-the-art NAS originally designed for computer vision, our method explores a $10^{12} \times$ larger and fine-grain search space, producing superior solutions, despite requiring less GPU memory and up to $14.22 \times$ lower search time.}
We test the proposed NAS on four real-world, edge-relevant tasks, involving audio and bio-signals: (i) PPG-based Heart-Rate Monitoring, (ii) ECG-based Arrythmia Detection, (iii) sEMG-based Hand-Gesture Recognition, and (iv) Keyword Spotting. 
Results show that, starting from a single seed network, our method is capable of obtaining a rich collection of Pareto optimal architectures, among which we obtain models with the same accuracy as the seed, and 15.9-152$\times$ fewer parameters. Moreover, the NAS finds solutions that Pareto-dominate state-of-the-art hand-tuned models for 3 out of the 4 benchmarks, and are Pareto-optimal on the fourth (sEMG).
\rev{Compared to three state-of-the-art NAS tools, ProxylessNAS, MorphNet and FBNetV2, our method explores a larger search space for TCNs (up to $10^{12} \times$) and obtains superior solutions, while requiring low GPU memory and search time.}
We deploy our NAS outputs on two distinct edge devices, the multicore GreenWaves Technology GAP8 IoT processor and the single-core STMicroelectronics STM32H7 microcontroller.
%
% , showing that they span two order-of-magnitude in terms of latency and energy consumption.
%
With respect to the state-of-the-art hand-tuned models, we reduce latency and energy of up to 5.5$\times$ and 3.8$\times$ on the two targets respectively, without any accuracy loss.

% %
% Furthermore, on the four benchmarks, we find solutions that Pareto-dominate the best hand-tuned state-of-the-art model.

% We also deploy the NAS outputs on two different embedded platforms in order to test the real performance of the obtained networks.
%
% On the first platform, the GAP8 System on Chip (SoC), \textcolor{red}{two}our best solutions achieve up to \textcolor{red}{XXXX times} model size compression, \textcolor{red}{5.5 times} latency reduction and \textcolor{red}{5.4 times} lower energy consumption \textcolor{blue}{MR: (su Dalia)} \textbf{DP: compared to what??}
% %
% On the second considered platform, the \textcolor{red}{STM???} MCU, we achieve up to \textcolor{red}{XXXX times} model size compression, \textcolor{red}{YYYY times} latency reduction and \textcolor{red}{ZZZZ times} lower energy consumption.
%

%% file: sections/01_Introduction.tex
\IEEEPARstart{D}{eep Learning} (DL) models are at the core of many time-series processing applications.
Notable examples are audio classification~\cite{dnn_speakerdetection_2018}, bio-signals analysis~\cite{temponet_2019, epilepsy_CNN_2018} and predictive maintenance~\cite{shm_dnn_review_2020,cerquitelli2021}.
For many years, the state-of-the-art DL models for time-series-based tasks have been Recurrent Neural Networks (RNNs)~\cite{rnn_review_2015}.
Recently, however, new architectures have been proposed as viable alternatives to RNNs, such as Attention-based Transformers and Temporal Convolutional Networks (TCNs)~\cite{tcn_original_2018}.
The latter, in particular, are uni-dimensional Convolutional Neural Networks (CNNs) specialized for time series, which have been shown to provide an accuracy comparable to RNNs, while providing computational advantages, namely higher arithmetic intensity, smaller memory footprint and more data reuse opportunities~\cite{tcn_original_2018}.
Thanks to these features, TCNs are particularly interesting for edge computing applications, where inference is directly executed on \revRemove{devices at the extreme edge of the Internet of Things (IoT)} \rev{Internet of Things (IoT) edge devices}, rather than on centralized servers in the cloud.
On-device inference requires small and efficient DL models, compatible with the limited memory spaces and tight energy budgets of edge nodes, while avoiding the transmission of raw data to the cloud provides many advantages, such as better privacy, higher energy efficiency and more predictable response latency~\cite{edge_computing_2016}.

To meet such tight constraints, however, selecting an efficient model such as a TCN is just the first step. Next, it is paramount to optimize its architectural hyperparameters based on the task at hand, so that the resulting network occupies as low memory and performs as few operations as possible to reach the desired accuracy level.
Nowadays, rather than manually, such architectural optimization is increasingly performed with automatic Neural Architecture Search (NAS) tools.
A plethora of different NAS approaches have been proposed in the last few years, and several of these works have targeted edge devices~\cite{nas_reinforcement_2016, fbnetv2_2020,morphnet_2018,mnasnet_2019,proxylessnas_2018}. However, to the best of our knowledge, none of them has focused specifically on models for time-series processing, nor specifically on TCNs, despite the unique features of these networks.

In fact, while TCNs share most of their key architectural features with standard CNNs, the peculiar 1D convolution operations at the heart of these networks increase the importance of some hyperparameters, such as the filters dilation and receptive field, resulting in a much larger variety in their values than what is common in 2D models for image-processing. Although there are NAS tools, originally designed for 2D CNNs, that can be easily extended to explore these parameters, they do so in a coarse-grain way, basically creating a different copy of all network layers for each architectural setting~\cite{proxylessnas_2018}. This approach results in highly memory- and time-consuming searches, requiring 100s of GPU hours even for relatively simple tasks, which in turn translate into large energy wastes and CO2 emissions. In contrast, lightweight NAS approaches explore a finer-grain space with lower complexity, but they achieve this result at the cost of focusing only on the key characteristics of a specific model type, e.g., the number of channels in each layer of 2D CNNs for computer vision~\cite{morphnet_2018,fbnetv2_2020}.

In Risso et al.~\cite{risso2021pit}, we proposed the first lightweight NAS explicitly designed for optimizing TCNs by tuning the dilation hyperparameter. In this work, we extend and complete~\cite{risso2021pit} by including the optimization of the receptive field and of the number of channels of all convolutional layers in a TCN, as well as the number of neurons in Fully Connected layers, with a search time comparable to that of a single, standard training. Starting from a single seed model, our proposed tool, called \textit{Pruning In Time (PIT)} can produce a rich set of Pareto optimal architectures in terms of number of operations/parameters versus accuracy.
The following are the main contributions of our work:
\begin{itemize}
    \item We frame the optimization of receptive field and dilation as a \textit{structured weight pruning}, in which additional trainable masking parameters are added to different layer's weights so that their binarized values encode valid settings of the architectural hyperparameters. These masks are then trained with a regularizer to reduce the model complexity as much as possible while preserving accuracy. While similar masking approaches already exist for optimizing the number of channels in a 2D convolutional layer~\cite{morphnet_2018}, our work is the first to extend this approach to filter size and dilation.
    \item We consider two different regularizers, targeting respectively the reduction of the number of parameters and of the number of inference operations. This allows us to enlarge and enrich the collection of Pareto architectures found by our NAS.
    \item  We test and validate PIT on four \revRemove{different }benchmarks relative to real-world time-series processing tasks where TCNs are commonly employed and for which a deployment on edge devices is relevant\revRemove{. The considered tasks are}: (i) PPG-Based Heart-Rate Monitoring; (ii) ECG-based Arrhythmia Detection; (iii) sEMG-based Hand-Gesture Recognition; (iv) Keyword Spotting. Results show that \revRemove{the proposed approach}\rev{PIT} can find multiple Pareto-optimal architectures starting from a single seed network, achieving 15.9-152$\times$ parameter reduction while maintaining the same accuracy of the seed.
    PIT is also capable of either matching or surpassing the accuracy and computational cost of state-of-the-art hand-tuned networks. 
    \revRemove{Furthermore, our approach Pareto-dominates ProxylessNAS, a popular NAS developed for computer vision, while exploring $10^{12}$ more architectures in $14.22\times$ lower time.}
    \rev{Furthermore, our approach Pareto-dominates three popular NAS tools developed for computer vision, thanks to the exploration of a larger search space.}
    \item  We deploy some of the relevant Pareto-optimal solutions found for each task on two different edge devices, in order to measure their memory footprint, latency and energy consumption. The two considered platforms are the multicore GAP8 IoT processor~\cite{flamand2018gap} and the single-core STM32H7 MCU~\cite{stm32h7}.
    The deployment results show that, at iso-accuracy, solutions found by PIT reduce energy consumption and latency up to 5.45$\times$ on GAP8 and up to 3.83$\times$ on the STM32H7, compared to hand-tuned networks.
\end{itemize}
The code of PIT is released as open-source at: \texttt{\href{https://github.com/EmbeddedML-EDAGroup/PIT}{https://github.com/EmbeddedML-EDAGroup/PIT}}.
The rest of the paper is structured as follows.
Section~\ref{sec:back_related} provides the required background and surveys some of the most relevant NAS methods proposed in the literature.
Section~\ref{sec:methods} presents the proposed methodology.
Section~\ref{sec:benchmark} details the target benchmarks while Section~\ref{sec:results} discussed the obtained results,
%
% along with ablation studies and comparisons with the state-of-the-art.
%
%Finally,
%
and Section~\ref{sec:conclusion} concludes the paper.
%
% \textbf{DP: questo bisogna capire dove metterlo. Per ora lasciamo in sospeso}
% \textcolor{red}{As shown in Sec.~\ref{subsec:abl_study} searching only for the dilation hyper-parameter is sub-optimal. On the other hand, giving more flexibility to the NAS enlarging the search space and giving multiple different knobs result in better exploration of the space of possible solutions.}
%

%% file: sections/02_Background.tex
\subsection{Temporal Convolutional Networks}
\label{subsec:tcn}
Temporal Convolutional Networks are 1-dimensional (1D) CNN variants that have recently gained significant traction for efficient time-series processing, obtaining state-of-the-art results in several tasks~\cite{risso2021robust,choi2019temporal,tsinganos2019improved}.
With respect to RNNs and their successive evolutions, such as the Long-Short Term Memory (LSTM) and Gated Recurrent Unit (GRU), TCNs are less affected by training-time issues, such as vanishing/exploding gradients and the large amount of training memory required by RNNs for long input sequences. Moreover, they also have computational advantages at inference time, since they share the better data locality and arithmetic intensity of standard CNNs, which makes them latency- and energy-efficient~\cite{tcn_original_2018}.

The main building blocks of TCNs are the same ones found in standard CNNs, i.e. Convolutional, Pooling and Fully Connected (FC) layers. However, the convolutional layers of a TCN are characterized by \textit{causality} and \textit{dilation}, two properties that make them suited for temporal inputs.

\emph{Causality} enforces that the outputs of convolutions do not violate the natural cause-effect ordering of events.
In practice, the outputs $y_{t}$ of a TCN convolution only depend on a finite set of past inputs $x_{[t-F;t]}$, where $t$ is a discrete index.
\emph{Dilation} is the mechanism used in TCNs for enlarging the receptive field of convolutions on the time axis, without requiring more trainable parameters and without increasing the number of operations required for inference.
It is a fixed step $d$ inserted between the input samples processed by each convolutional filter.
Eq.~\ref{eq:1d_conv} summarizes the 1D dilated convolution operation implemented by TCN layers:
\begin{equation}\label{eq:1d_conv}
y_t^m = \sum_{i=0}^{K-1} \sum_{l=0}^{C_{in}-1} x_{ts-d\,i}^l \cdot W_i^{l,m}\ , \forall m \in [0,C_{out} - 1],\forall t \in [0, T-1]
\end{equation}
where $x$ and $y$ are the input/output activations, $T$ is the output sequence length, $W$ the array of filter weights, $C_{in}$/$C_{out}$ the number of input/output channels, $K$ the filter size and $s$ the stride. We also define $F = d\cdot (K-1) + 1$ the \textit{receptive field} of the layer.
\revFloatRemove{
\begin{figure}[t]
  \centering
  \includegraphics[width=.9\columnwidth]{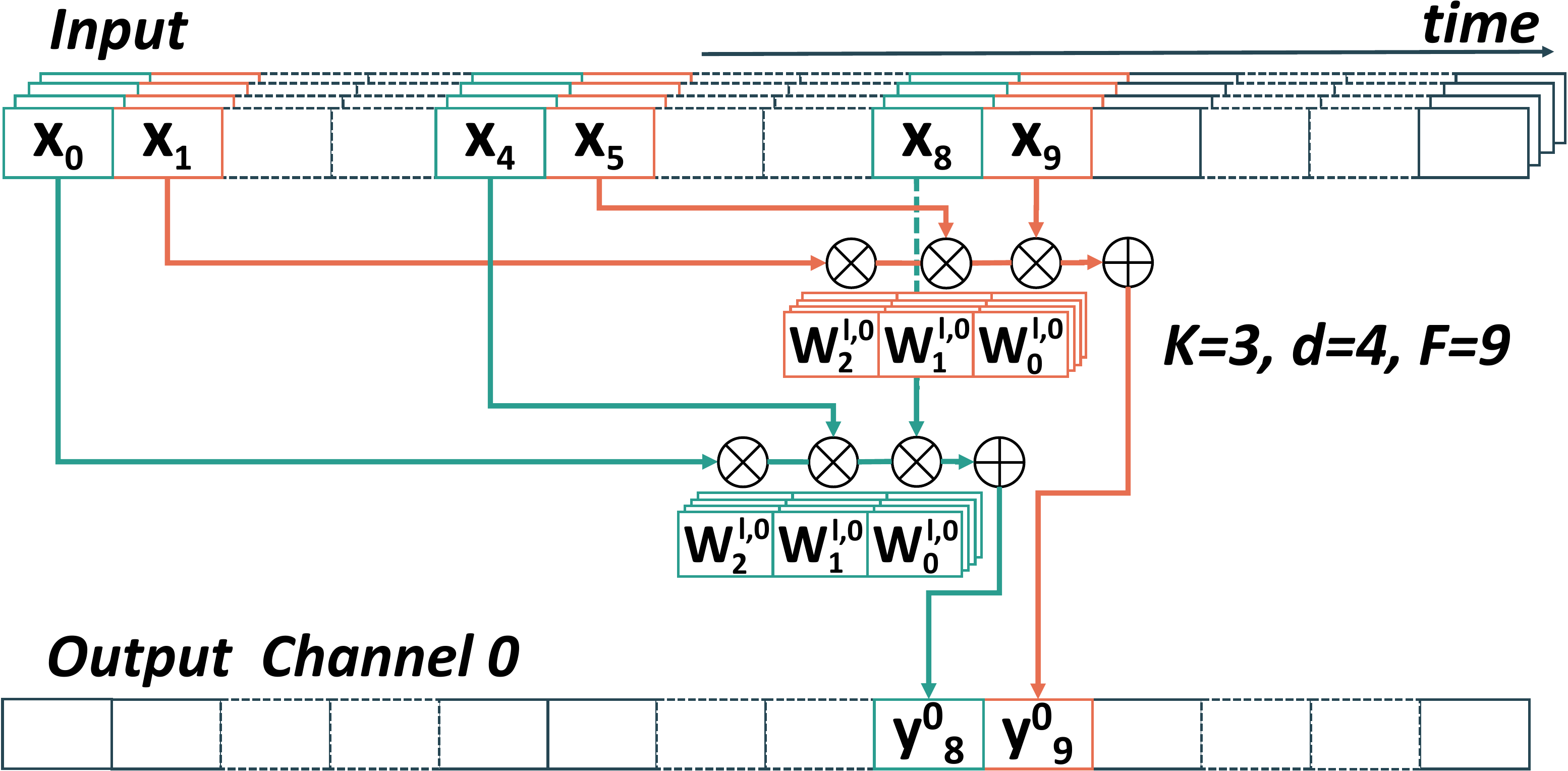}
  %\vspace{-0.4cm}
  \caption{\revRemove{Generation of the first 2 output time samples in a TCN layer with $K$ = 3, $d$ = 4, $F$ = 9 and $s$ = 1 for the first output channel ($m$ = 0).}}
  \label{fig:tcn_layer}
\end{figure}
}
\revRemove{Figure~\ref{fig:tcn_layer} depicts a TCN Convolutional layer visually.
Originally, TCNs were proposed in~\cite{tcn_original_2018} as fully convolutional networks with residual connections.
Modern TCNs include also other layers such as strided convolutions, pooling and fully-connected~\cite{temponet_2019}.}
\subsection{Neural Architecture Search}
\label{subsec: nas}
In recent years, several manually designed efficient and compact convolutional neural network architectures for edge devices have been proposed, including early MobileNets~\cite{sandler2018mobilenetv2}, ShuffleNets~\cite{ma2018shufflenet}, EfficientNet~\cite{efficientnet}, SqueezeNet~\cite{SqueezeNet}, etc.
While these models are very efficient, obtaining them required a long and time-consuming manual tuning of hyper-parameters, which has to be repeated from scratch when considering a different target task, or a different deployment target.
% 
% For example, MobileNets, a class of networks explicitly designed for smartphone devices, are significantly too large to be deployed on extreme-edge devices such as Microcontrollers (MCUs).

To solve this issue, many automated or semi-automated methods to optimize neural network architectures, easing the burden of designers, have been proposed.
These approaches, generally denoted as Neural Architecture Search (NAS) algorithms, explore a large design space made of different combinations of layers and/or hyper-parameter values, selecting solutions that optimize a cost metric. The latter is often a function of both the accuracy of the network, and its computational cost (e.g., number of parameters or inference operations).

\rev{Table~\ref{tab:sota} qualitatively compares some of the most relevant works in this field, in terms of search time, memory requirements during training (Mem.), search space size, and possibility to vary the topology (number and type of layers) of the resulting NNs. For Time and Mem., smaller is better, whereas for Search Space, larger is better.}
Early NAS tools were based on Reinforcement Learning (RL) \cite{nas_reinforcement_2016, mnasnet_2019, NASNet, Baker2017DesigningNN} or Evolutionary Algorithms (EA) \cite{real2017large}.
At each search iteration, these methods sample one or more architectures from the search space.
Sampled networks are then trained to convergence to evaluate their accuracy (and possibly cost), which is then used to drive the next sampling.
%
% used to tune the sampling.process of EA or is fed back to the RL agent.
%
The repeated training in each iteration is the main drawback of these tools, for which a single search requires 1000s of GPU hours, even on relatively simple tasks. \rev{Accordingly, these methods are associated with large search time in Table~\ref{tab:sota}. Memory occupation is low and comparable to a standard training, since each sampled architecture can be trained separately. The search space size is virtually unlimited, and these tools can easily support variable topologies. Notable exceptions are \cite{nas_reinforcement_2016}, which searches over a fixed convolutional topology of a variable number of layers without varying their type and the connections between them, and \cite{NASNet}, that constrains its search space to a set of only 13 different layers per node.} 

\rev{To solve the search time issue of RL and EA methods, more recent \textit{Differentiable} NAS (DNAS) approaches have proposed the so-called \textit{supernets}~\cite{darts_2019}.}
Supernets are DNNs that include all possible alternative layers to be considered during the optimization. For instance, a single supernet layer might include multiple Convolutional layers with different kernel sizes, operating in parallel.
The problem of choosing a specific architecture is then translated into the problem of choosing a \textit{path} in the supernet~\cite{darts_2019}.
The choice between the different paths is encoded with binary variables, jointly trained with the standard weights of the network using gradient-based learning.
\revRemove{In particular, t}\rev{T}o search for accurate and efficient architectures, DNAS tools enhance the normal training loss function with an additional differentiable regularization term that encodes the cost of the network. Typical cost metrics are the number of parameters and the number of Floating Point Operations (FLOPs) per inference~\cite{morphnet_2018}. \revRemove{In mathematical terms,}\rev{Mathematically,} DNAS tools search for:
\begin{equation}\label{eq:dnas}
\min_{W, \theta} \mathcal{L}(W; \theta) + \lambda \mathcal{R}(\theta)
\end{equation}
where $\mathcal{L}$ is the standard loss function, $W$ is the set of standard trainable weights (e.g., convolutional filters), $\theta$ is the set of additional NAS-specific trainable parameters that encode the different paths in the supernet, $\mathcal{R}$ is the regularization loss that measures the cost of the network and $\lambda$ is a hand-tuned \textit{regularization strength}, used to balance the two loss terms.

\input{tables/00_table_NAS_compressed}

While DNAS algorithms are more efficient than early RL/EA-based solutions, training the entire supernet still requires huge computational resources both in terms of \rev{training time and memory occupation. This, in turn, translates in a reduction of the explored search space for practical DNASes such as~\cite{darts_2019}, which have to limit the search to few alternatives per layer, in order to keep the memory occupation under reasonable bounds.} 
\revRemove{Therefore, t}The authors of \cite{proxylessnas_2018} have proposed ProxylessNAS, an advanced DNAS that reduces the memory requirements, keeping in memory at most two supernet paths for each batch of inputs.
%
% In particular, ProxylessNAS couples each layer alternative in the supernet with a trainable architecture parameter.
%
In ProxylessNAS, the normal weights and the additional parameters encoding supernet paths are trained and updated in an alternate manner. 
First, path parameters are frozen, and based on their current value, one sub-architecture of the supernet is stochastically sampled. Then, the weights of the sampled architecture are updated based on the training set.
Second, the normal weights are frozen and the architectural parameters are trained on the validation set.
This second phase updates two different paths at a time, sampling them from a multinomial distribution. \rev{In turn, this clever strategy allows ProxylessNAS to explore a significantly larger search space compared to other DNAS tools.}

A further evolution in the direction of lightweight NAS is constituted by DMaskingNAS \cite{fbnetv2_2020}, fine-grain NAS~\cite{morphnet_2018} and Single-Path NAS~\cite{stamoulis2020single} approaches.
In these solutions, the supernet is replaced by a single, usually large, architecture with a unique path. Optimized architectures are found as modifications of this initial \textit{seed model}, obtained tuning hyper-parameters, such as the number of channels in each layer~\cite{morphnet_2018}.
The key mechanism that enables this tuning within a normal training loop is the use of \textit{trainable masks}, used to prune parts of the network.
DMaskingNAS tools pursue the same DNAS objective of (\ref{eq:dnas}), where $\theta$ now represents the set of trainable masks.
FBNet-V2 \cite{fbnetv2_2020}, for instance, uses a set of dedicated masks, each of which encodes a different number of output channels or a different spatial resolution, and is weighted with a trainable parameter.
At the end of the search, the mask coupled with the largest parameter is used to determine the final architectural setting.
Similarly, MorphNet \cite{morphnet_2018} exploit as masking parameters the  pre-existing multiplicative terms of batch normalization layers \cite{ioffe2015batch}. When these parameters assume a value lower than a threshold, the corresponding channels/feature maps from the preceding Convolutional layer are eliminated.

\revRemove{In general, the search space of these approaches is slightly more constrained with respect to supernet-based ones. For example, they do not allow to select between alternative layers (e.g., standard convolution versus depth-wise + point-wise).}
\rev{These approaches are more constrained than supernet-based ones in terms of NN topology. In fact, they do not allow to select between alternative layers (e.g., standard convolution versus depth-wise + point-wise convolution).}
On the other hand, they have two key advantages. First, they have much lower memory cost and search time, while still being able to find high-quality architectures. Crucially, the search time of a DMaskingNAS is comparable to a standard network training.
\rev{Second, some DMaskingNASes (including our work) can explore the search space at a much finer grain}. For example, MorphNet~\cite{morphnet_2018} can easily select between 1 and 32 output channels in a Convolutional layer with a granularity of 1, by starting from a 32 channels seed layer, and eliminating those corresponding to the smallest batch normalization multiplicative parameters. Obtaining the same result with a standard DNAS would require a very large supernet, with 32 parallel convolutional layers. The masking and super-net approaches can also be combined, to bypass the limitations of DMaskingNAS\revRemove{, as shown in}~\cite{Chaudhuri2020}.

The NAS literature referenced above focuses almost exclusively on 2D-CNNs for computer vision.
None of the existing approaches has been applied to time-series processing tasks, despite the fact that a large amount of edge-relevant real-world tasks deal with uni-dimensional time-dependent signals (e.g., bio-signals, audio, energy-traces, sensor readings from industrial machines, etc).
\rev{
Our work tries to fill this gap, by proposing a novel DMaskingNAS that targets the optimization of 1D networks. Moreover, the working principles of our tool are general, and could form the basis for a more general NAS, able to explore temporal hyper-parameters of arbitrary N-dimensional Convolutional layers (e.g., including also 3D-CNNs for spatio-temporal data processing), although this paper focuses exclusively on TCNs.}

%% file: tables/00_table_NAS_compressed.tex
\begin{table}[t]
\centering
\caption{\rev{State-of-the-art NAS (Values: $\uparrow =$ large, $\nearrow =$ medium, $\downarrow =$ small).
}}
\label{tab:sota}
\renewcommand{\arraystretch}{1.1}
\footnotesize
\rev{
\begin{tabular}{|l|cccc|}
\hline
         & Time & Mem. & Search Space & Topology \\ \hline\hline
\multicolumn{5}{|l|}{\textbf{Reinforcement Learning} } \\ \hline\hline
Zoph et al. \cite{nas_reinforcement_2016}                & $\uparrow$        & $\downarrow$               & $\nearrow$        & Variable$^*$ \\ \hline
MNASNET \cite{mnasnet_2019}                  & $\uparrow$         & $\downarrow$               & $\uparrow$          & Variable  \\ \hline
NASNET \cite{NASNet}                          & $\uparrow$         & $\downarrow$               & $\nearrow$        & Variable  \\ \hline
MetaQNN \cite{Baker2017DesigningNN}          & $\uparrow$         & $\downarrow$               & $\uparrow$          & Variable  \\ \hline\hline
\multicolumn{5}{|l|}{\textbf{Evolutionary} } \\ \hline\hline
Real et al. \cite{real2017large}              & $\uparrow$         & $\downarrow$               & $\uparrow$         & Variable  \\ \hline \hline
\multicolumn{5}{|l|}{\textbf{DifferentiableNAS}}  \\ \hline\hline
DARTS \cite{darts_2019}                           & $\nearrow$       & $\uparrow$             & $\downarrow$         & Variable  \\ \hline
ProxylessNAS \cite{proxylessnas_2018}                      & $\nearrow$       & $\nearrow$             & $\nearrow$  & Variable  \\ \hline \hline
\multicolumn{5}{|l|}{\textbf{DmaskingNAS} } \\ \hline\hline
FBNetV2 \cite{fbnetv2_2020}                             & $\downarrow$         & $\downarrow$               & $\uparrow$        & Fixed    \\ \hline
MorphNet \cite{morphnet_2018}                           & $\downarrow$         & $\downarrow$               & $\nearrow$          & Fixed    \\ \hline
S.-Path NAS   \cite{stamoulis2020single}            & $\downarrow$         & $\downarrow$               & $\nearrow$          & Fixed    \\ \hline 
\textbf{PIT (this work)}                                                     & $\downarrow$          & $\downarrow$               & $\uparrow$        & Fixed \\ \hline
\multicolumn{5}{l}{$^*$ Depth only } \\

\end{tabular}
}
\end{table}

%% file: sections/03_Method.tex
We name our proposed tool \textit{Pruning in Time (PIT)}, since it targets networks that process time-series, and the core mechanism of a DMaskingNAS is very similar to structured pruning~\cite{yu2017scalpel}.
PIT explores the architectures of convolutional and fully-connected (FC) layers, the two most compute- and memory-expensive operations present in TCNs. For each convolutional layer, PIT
jointly explores the \textit{number of channels ($C_{out}$)}, the \textit{receptive field ($F$)}, and the \textit{dilation ($d$)}. 
Moreover, by tuning both $F$ and $d$, it also indirectly affects the \textit{filter size} $K$.
To the best of our knowledge, no DMaskingNAS from literature has optimized the receptive field or the dilation, even for 2D-CNNs.
%
% tunes the number of channels in each layer.
Similarly, PIT can also optimize the number of output neurons of FC layers\footnote{This can be seen as a corner case of the $C_{out}$ optimization, since FC layers are just a special case of 1D convolutions with $F=K=d=1$ and $C_{out}$ equal to the number of output neurons. 
Accordingly, the rest of this section describes PIT's functionality for convolutions.}.

We provide an overview of the search space explored by our tool and of its general working principle in Section~\ref{subsec:search_space}.
Then, we detail the mechanisms used to generate differentiable masks for each considered hyper-parameter in Sections~\ref{subsec:ch_search}-\ref{subsec:joint_search}.
%
% Note that this is a key feature of our algorithm, allowing it to learn the mask values with gradient-based optimizers and therefore be faster than multi-path DNAS \cite{proxylessnas_2018}.
%
Finally, the two cost regularizers used to drive the search and the overall training procedure are described in Section~\ref{subsec:regularizer} and \ref{subsec:train_proc} respectively.
Table~\ref{tab:notation} summarizes the main mathematical symbols used throughout the paper.
\input{tables/notation_recap}
\subsection{Search Space} \label{subsec:search_space}
\begin{figure}[t]
  \centering
  \includegraphics[width=.9\columnwidth]{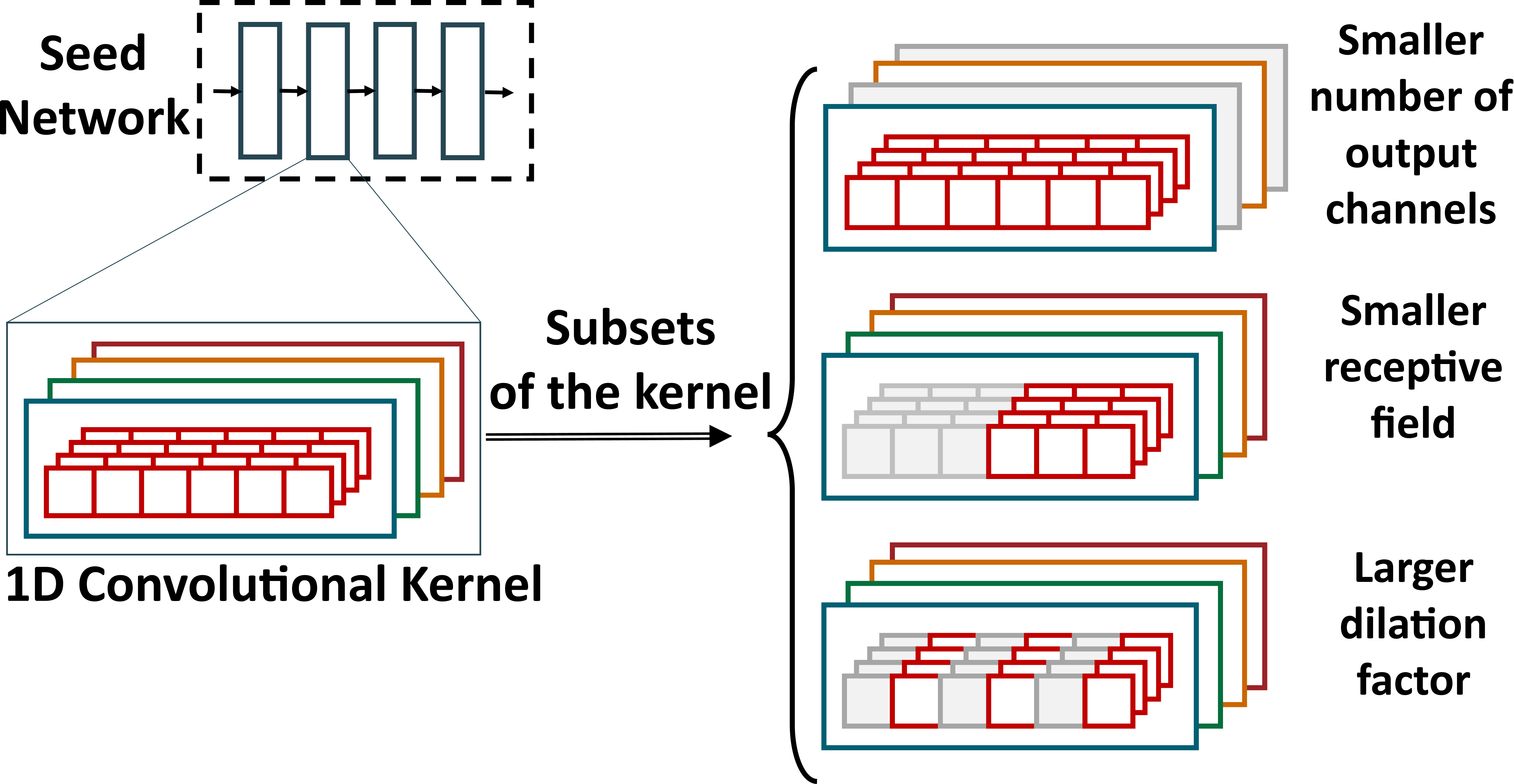}
  %\vspace{-0.4cm}
  \caption{Search space of PIT.}
  \label{fig:search_space}
\end{figure}

As shown in Figure~\ref{fig:search_space}, PIT's search space encompasses all sub-architectures derived from a seed TCN by tweaking the three aforementioned hyperparameters. In particular, PIT can decide to \textit{reduce} $C_{out}$ or $F$, and to \textit{increase} $d$ with respect to the seed, all of which have the effect of reducing the complexity and memory occupation of the layer.
%
% The search is performed leveraging an additional set of trainable mask parameters.
%

To achieve this objective, each convolutional/FC layer of the seed is modified to become a function $L_{n}(W^{(n)};\theta^{(n)})$ of its original weights tensor $W^{(n)}$ and of a new set of architectural parameters $\theta^{(n)}$.
For a TCN with $N$ layers, the search space of PIT is therefore defined by the set:
\begin{equation}\label{eq:search_space}
\mathcal{S} = \{ L_{n}(W^{(n)};\theta^{(n)}) \}_{n=0}^{N-1}
\end{equation}

During the search, the elements of $\theta^{(n)}$ are properly combined to form a \textit{binary mask} $\Theta^{(n)}$, which is used to \textit{prune} a portion of the layer weights.
In practice, an architecture $\hat{\mathcal{S}}$ is sampled from $\mathcal{S}$ in each search iteration, by performing the Hadamard product between $W^{(n)}$ and $\Theta^{(n)}$, i.e., $\hat{\mathcal{S}} = \{ L_{n}(W^{(n)} \odot \Theta^{(n)}) \}_{n=0}^{N-1}$.
This eliminates the portions of $W^{(n)}$ that correspond to 0-valued mask elements, effectively letting the seed layer produce the same output that would be obtained with a smaller number of channels or receptive field, or with a larger dilation.
The way in which $\Theta^{(n)}$ is generated from $\theta^{(n)}$ to produce this effect is the topic of Sections~\ref{subsec:ch_search}-\ref{subsec:dil_search}.

Having \textit{binary} masks
%
% is critical for the correct functionality of the tool.
%
is required to either completely eliminate slices of $W^{(n)}$ (with value 0) or keep them untouched (with value 1) when sampling an architecture with the Hadamard product. In practice, this corresponds to sampling only \textit{feasible} architectures (with integer $C_{out}$, $F$ and $d$).
To this end, we
%
% transform the floating point mask $\hat{\bm{\Theta}}^{(n)}$ into its binary version $\bm{\Theta}^{(n)}$
%
binarize $\Theta^{(n)}$
in the forward-pass of our search/training, applying an Heaviside step function with a fixed threshold $\mathit{th} = 0.5$. 
%
% \begin{equation} \label{eq:binarization}
% \mathcal{H}(\hat{\theta} - \delta) = 
% \begin{cases}
%   1, & \text{for } \hat{\theta} \ge \delta  \\    
%   0, & \text{for } \hat{\theta} < \delta
% \end{cases}
% \end{equation}

At the same time, we also need to make the $\theta^{(n)} \rightarrow \Theta^{(n)}$ transformation differentiable, in order to embed the search into the standard gradient-based training of the network, learning contextually both the weights $W^{(n)}$ and the architectural parameters $\theta^{(n)}$.
To cope with the Heaviside function derivation issues, i.e., derivative equal to 0 almost everywhere and not existent in $\delta$, we follow the approach proposed in BinaryConnect \cite{courbariaux2016binarized}, based on a Straight-Through Estimator (STE). 
Accordingly, during the backward-pass, the step function is simply replaced with an identity.
%

%The different hyper-parameters explored in each layer $L_{n}$ by tweaking the $\hat{\theta}^{(n)} \in \hat{\Theta}^{(n)}$ parameters during learning are depicted in the left-most part of Fig.~\ref{fig:search_space} with the corresponding representation in terms of kernel subsets.
%
%The three knobs that we can act on to prune out parts of the kernel tensors are: 
%
%\begin{itemize}
%    \item[(i)] Number of output channels.
%    \item[(ii)] Kernel size.
%    \item[(iii)] Dilation Factor.
%\end{itemize}
%

For notation simplicity, in the rest of the section we divide $\theta^{(n)}$ parameters in three groups: $\alpha^{(n)}$, used to tune the number of channels, $\beta^{(n)}$, which tune the receptive field, and $\gamma^{(n)}$, which affect the dilation factor. We also drop the superscript $(n)$ when not needed.
In PIT, each of these three groups of parameters is used to generate an \textit{independent} binary mask, which can be then combined with the other two.
%
% We describe this process in the following sections.
%
% Importantly,
%
Having independent masks for $C_{out}$, $F$ and $d$, gives PIT the flexibility to optimize the three hyper-parameters either separately or jointly.
%
% Moreover, the way masks are generated makes PIT flexible also in terms of the \textit{granularity} of the search, i.e., of the possible values of $C_{out}$, $F$ and $d$ that can be obtained when sampling an architecture.
%
At most, during a joint search, PIT explores:
\begin{equation}\label{eq:search_space_size}
|\mathcal{S}| \approx \prod_{n=0}^{N-1} (C_{out,seed}^{(n)} \cdot F_{seed}^{(n)} \cdot \lceil\text{log}_2(F_{seed}^{(n)})\rceil)
\end{equation}
different solutions, where $C_{out,seed}$ and $F_{seed}$ in (\ref{eq:search_space_size}) are those of the seed layers. The logarithmic term in (\ref{eq:search_space_size}) comes from the fact that we only consider power-of-2 dilation factors, as detailed in Section~\ref{subsec:dil_search}.
For a relatively small seed with  $N = 8$, $F_{seed}^{(n)} = 17$, and $C_{out,seed}^{(n)} = 128\ \forall n$, this corresponds to evaluating $\approx 10^{32}$ architectures in a single training.

\subsubsection{Channels Search} \label{subsec:ch_search}

\begin{figure}[ht]
  \centering
  \includegraphics[width=1.\columnwidth]{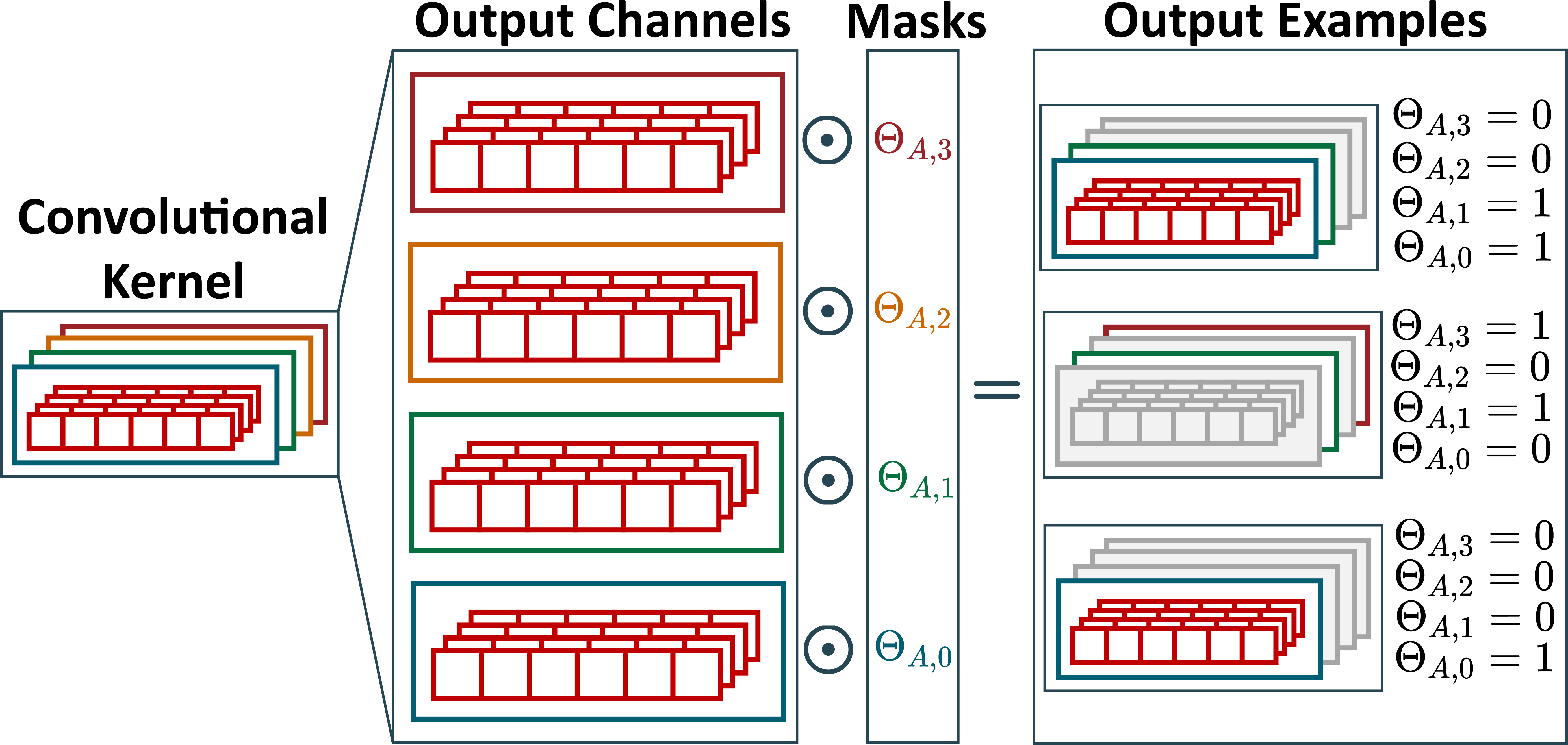}
  %\vspace{-0.4cm}
  \caption{Channels search example. Each $\Theta_{A,m} = 0$ zeroes-out the $m$-th convolutional filter, i.e., a slice of size $K \times C_{in}$ of the weights tensor $W$.}
  \label{fig:ch_search}
\end{figure}

To explore the number of channels in each convolutional layer, we take inspiration from~\cite{morphnet_2018}. In that work, the parameters of batch normalization (BN) layers~\cite{ioffe2015batch} were transformed into binary masks to prune entire output channels and explore the space of all sub-layers with $C_{out} < C_{out,seed}$. \revRemove{Indeed, when a BN layer follows a convolutional one, each output channel is obtained as:}
\revFloatRemove{
\begin{equation}\label{eq:conv+BN}
\revEqRemove{\Tilde{y}_t^m = \gamma^m \cdot y_t^m}
\end{equation}
}
\revRemove{
where $y_t^m$ is the output of (\ref{eq:1d_conv}) and $\gamma^m$ the multiplicative factor of BN.
When the latter is binarized to 0, the entire $m$-th output channel is effectively pruned.
}
However, requiring the presence of a BN layer after each convolution, although common in modern 2D-CNNs, still limits the applicability of the approach of~\cite{morphnet_2018}.
%
% This is especially true for very small edge TCNs, for which even the relatively low computational cost of BN is non-negligible. 
%
\rev{Therefore, in PIT, we decouple the channel search from BN, and instead we exploit a dedicated trainable set of parameters $\alpha$ to zero-out entire filters from the $W$ tensor of convolutional layers.}
PIT treats each output channel independently. So, it uses an $\alpha$ array of length $C_{out,seed}$, and it generates binary masks simply as:
\begin{equation}\label{eq:alpha_mask}
\Theta_{A} = \mathcal{H}(|\alpha|)
\end{equation}
where $\mathcal{H}$ is the Heaviside binarization.
Then, the layer function defined in (\ref{eq:1d_conv}) is modified to: 
\begin{equation} \label{eq:mask_ch_conv}
    \Tilde{y}_{t}^{m} = \sum_{i=0}^{K-1} \sum_{l=0}^{C_{in}-1} x_{ts-di}^l \cdot (\Theta_{A,m} \cdot W_i^{l,m})
\end{equation}
\rev{In practice, each binarized mask element is multiplied with all the weights of the \textit{same convolutional filter}, i.e., with an entire slice of the weights tensor over the output channels axis. Each filter multiplied with a 0-mask effectively removes the corresponding output channel from the layer.}
Figure~\ref{fig:ch_search} depicts the application of $\Theta_{A}$ parameters to a simple layer with $C_{out,seed} = 4$. 

% Alternatively, we can also easily explore $C_{out}$ with a granularity $gr_\alpha > 1$. In this case, groups of $gr_\alpha$ consecutive channels are pruned together. Then, the length of the $\alpha$ vector becomes $\left\lfloor \frac{C_{out}}{gr_\alpha} \right\rfloor$ and the binary mask is created as: $\Theta_{A,m} = \mathcal{H}(\alpha_{\lfloor m/gr_\alpha \rfloor})$. In general, we can express the differentiable function that generates binary channel masks in matrix form as:
% \begin{equation}
%     \Theta_A = C_\alpha \cdot \alpha
% \end{equation}
% %
% where $C_{\alpha}$ is a constant matrix of 0s and 1s that can be generated procedurally based on the values of $C_{out}$ and $gr_\alpha$

%

Noteworthy, besides reducing the number of channels, PIT can also \textit{eliminate} entire layers from the network, if the latter includes skip-connections.
In particular, if all the $\Theta_{A,m}$ of a convolutional layer are zeroed-out, then the inputs only flow through the skip connection, effectively reducing the number of the layers in the network by one. If skip connections are not present, instead, at least one output channel is always kept active to avoid breaking the network connectivity.

\rev{Our channels search scheme differs significantly from existing DMaskingNAS such as FBNetV2~\cite{fbnetv2_2020}, since we mask \textit{weights tensors} rather than output activations. Fundamentally, as explained below, this makes our method easily extensible to the exploration of other hyper-parameters such as $F$ and $d$, which would be much more difficult to optimize with an activations mask. Moreover, we use \textit{independent} binarized parameters to mask each channel, rather than a set of predefined masks with an increasing number of trailing 0s, combined via Gumbel Softmax, as done in~\cite{fbnetv2_2020}. This, in turn, means that we can eliminate \textit{any} combination of channels, not just the trailing ones.}

\subsubsection{Receptive Field Search} \label{subsec:rf_search}
The second critical hyperparameter that we explore is the receptive field $F$, i.e., the
range of input time-steps involved in a convolution.
In standard convolutions, the receptive field is equal to the filter size ($F = K$). However, as detailed in Section~\ref{subsec:tcn}, this no longer holds for TCNs when the dilation factor $d$ is $>1$, and the general relation becomes:
$F = (K - 1) \cdot d + 1$.
As noted at the beginning of this section, by exploring both $F$ and $d$, PIT also indirectly optimizes the filter dimension $K$.

The receptive field is explored using an array of additional trainable parameters $\beta$ of length $F_{seed}$.
%
% $len(\beta) = \left \lfloor \frac{F}{\mathit{gr}} \right \rfloor$
% %
% where $F$ is the seed receptive field and $gr \ge 1$ is the granularity of the exploration. Having a granularity greater than 1 for the receptive field allows, for instance, to force  network to only use even/odd receptive fields. \textbf{@MR: justify!!}
%
Differently from the output channels, however, the $\beta$ need to be further combined to define the corresponding binary differentiable masks. The reason is that,
%
% setting some $\beta$ to 0 should generate the same output that would be produced by a layer with a smaller $F$. As shown in Figure~\ref{fig:rf_search}, in order to obtain this effect,
%
to ``simulate'' the effect of a smaller receptive field through masking,
it is not sufficient to mask \textit{any} set of time-slices in the weights tensor: in a causal TCN convolution, the receptive field extends exclusively in the past. Thus, the slices that should be eliminated are always the ``oldest'' ones, i.e., those that are multiplied with input time-steps that are farthest in the past.
To do so, we derive elements of the binary masks $\Theta_{B}$ from $\beta$ as:
\begin{equation} \label{eq:beta_aggregation}
    \Theta_{B,i} = \mathcal{H} \left( \sum_{j = 1}^{F_{seed} - i} |\beta_{F_{seed} - j}| \right)
\end{equation}

\begin{figure}[ht]
  \centering
  \includegraphics[width=1.\columnwidth]{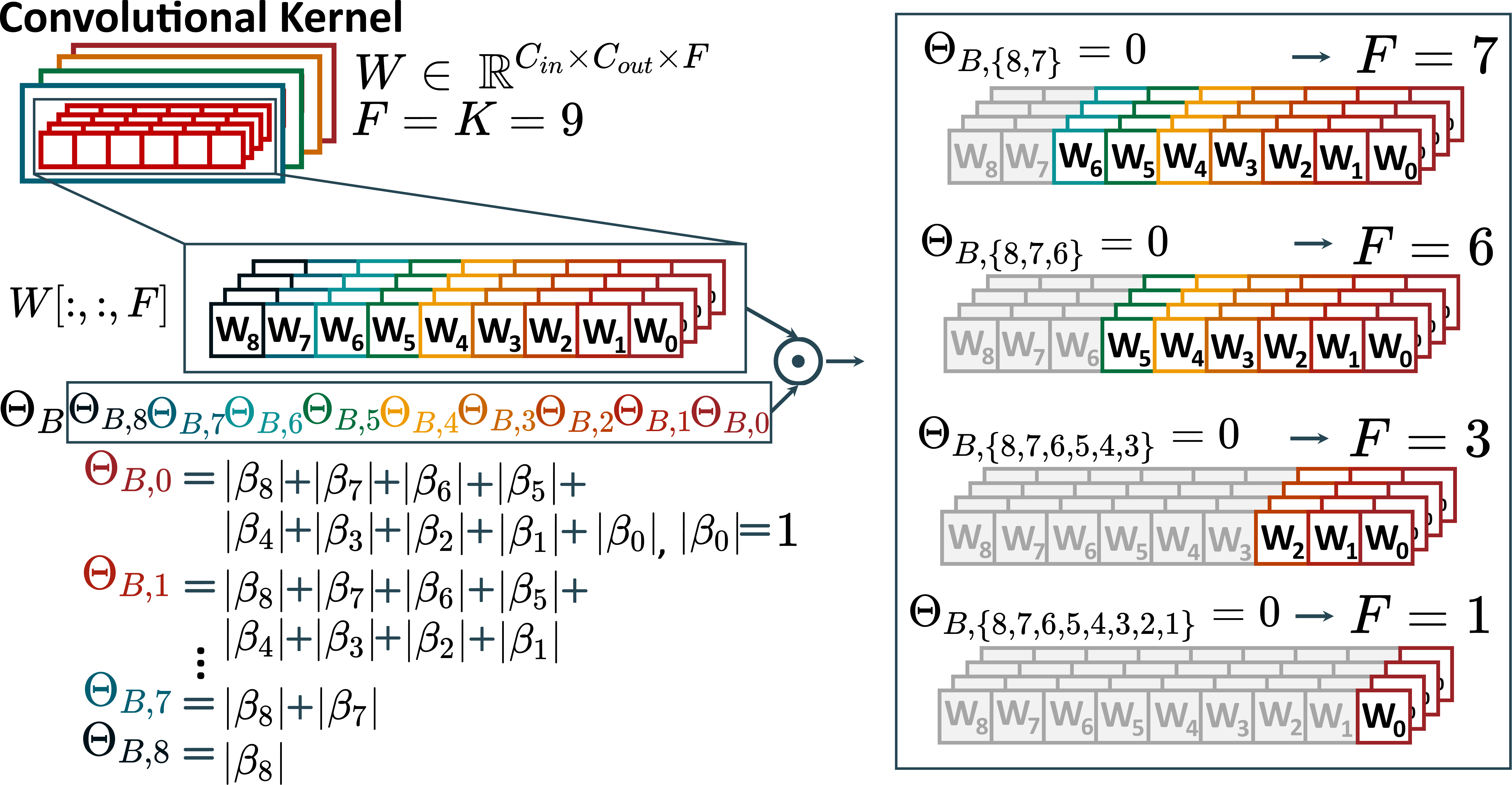}
  %\vspace{-0.4cm}
  \caption{Receptive field search example. Each $\Theta_{B,i} = 0$ eliminates the contribution of 1 input time-step from the convolution output, by zeroing out a time-slice of size $C_{out} \times C_{in}$ of the weights tensor $W$.}
  \label{fig:rf_search}
\end{figure}

Each $\Theta_{B,i}$ is then multiplied with a time-slice of the $W$ tensor during the forward-pass, as shown in
Figure~\ref{fig:rf_search}.
Therefore, when searching for the receptive field, (\ref{eq:1d_conv}) becomes:
\begin{equation} \label{eq:mask_rf_conv}
    y_t^m = \sum_{i=0}^{K-1} \sum_{l=0}^{C_{in}-1} x_{ts-di}^l \cdot (\Theta_{B,di} \odot W_i^{l,m})
\end{equation}
%
% where $d$ has been set to 1, since this is what PIT does to the seed network when searching for both receptive field and dilation, as explained later.
%
Thanks to the construction of (\ref{eq:beta_aggregation}), we have that if $i > j$, then $ \Theta_{B,i} \le \Theta_{B,j}$. In turn, this ensures that the first weight slices to be pruned are always the leftmost ones,  as shown in the example on the right of Figure~\ref{fig:rf_search}.
Importantly, $\beta_0$ is always kept constant and equal to 1. This ensures that, once binarized, $\Theta_{B,0}$ is also always $=1$, and consequently, that all convolutions take \textit{at least one time-step} as input.

\begin{figure}[t]
  \centering
  \includegraphics[width=.8\columnwidth]{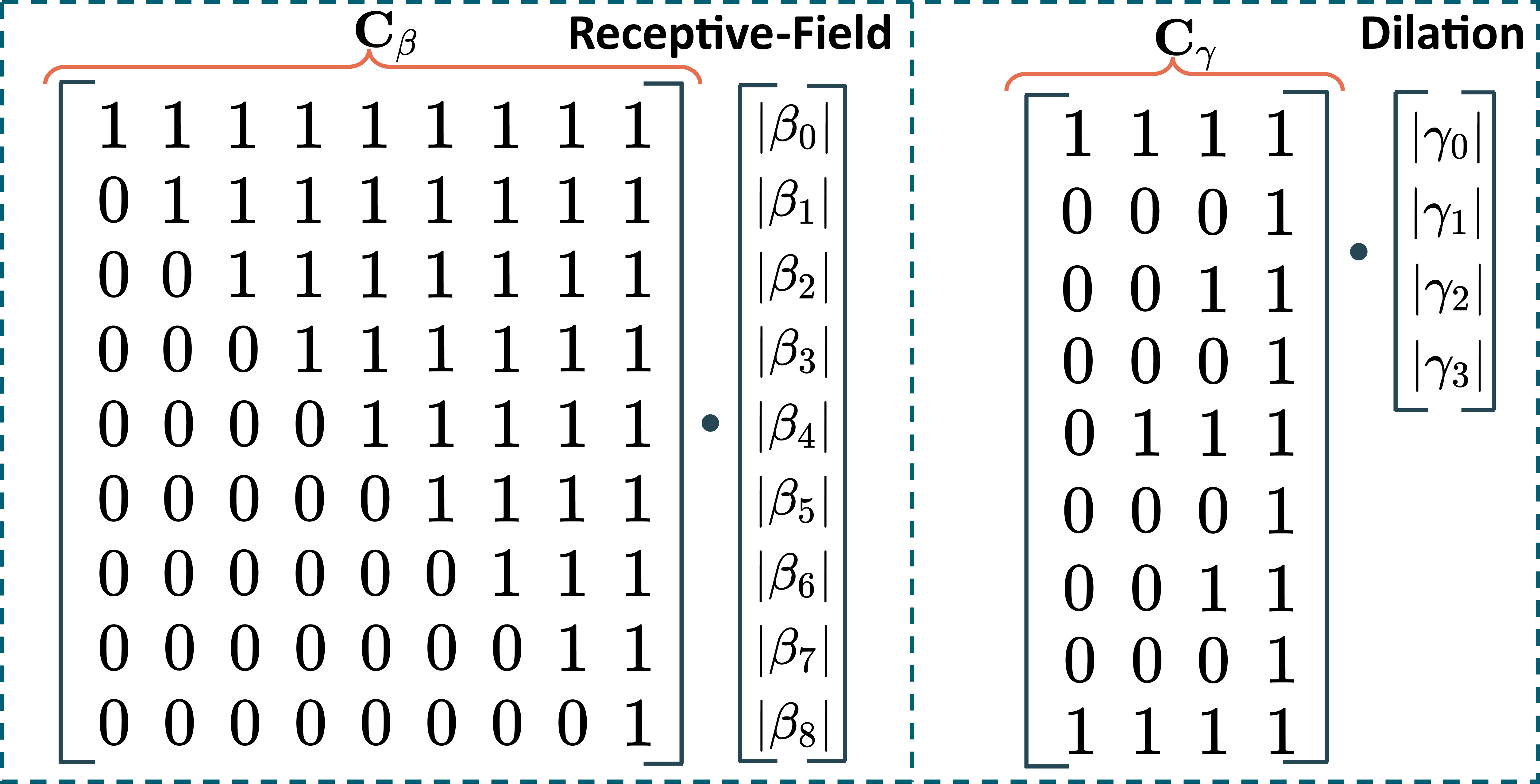}
  %\vspace{-0.4cm}
  \caption{Example of conversion between trainable architectural parameters $\beta$ and $\gamma$ and corresponding binary masks $\Theta_B$ and $\Theta_\Gamma$, for a layer with $F_{seed} = 9$.}
  \label{fig:matrix_transf}
\end{figure}

In practice, for efficiency reasons, we generate binary masks using the matrix transformation:
\begin{equation} \label{eq:beta_mask}
    \Theta_{B} = \mathcal{H} \left( C_{\beta} \cdot | \beta |\right)
\end{equation}
where $C_{\beta}$ is a constant upper triangular matrix of 1s generated once at the beginning of a search, as shown on the left of Figure~\ref{fig:matrix_transf}.
%
% $d_{\beta}$ is composed by $F-1$ 0s and a final element equal to 1.
% %
% Also, $C_{\beta}$ present size $len(\beta) \times F$ and each column from $0$ to $F-1$ is generated using the relation: $\mathit{col}[i:(\mathit{gr} \cdot i+1)] = \#(len(\beta - 1 - \mathit{gr} \cdot i)) \mathit{leading} \: 0s \: \text{and} \: \#(1 + \mathit{gr} \cdot i) \mathit{trailing} \: 1s$.
% %
% Conversely, the last column of $C_{\beta}$ is composed by all 0s.
%
% \begin{figure}[ht]
%   \centering
%   \includegraphics[width=1.\columnwidth]{figures/rf_mask.pdf}
%   %\vspace{-0.4cm}
%   \caption{.}
%   \label{fig:rf_mask}
% \end{figure}
% %
% Fig.~\ref{fig:rf_mask} provides an example of $d_{\beta}$ and $C_{\beta}$ and the entire transformation of Eq.~\ref{eq:beta_mask} for the case $F = 9$.
% %
%

\subsubsection{Dilation Search} \label{subsec:dil_search}

Lastly, PIT also explores the dilation factor $d$. 
Similarly to the receptive field, also searching for dilation imposes some constraints on the portions of the weights tensor that should be pruned by our NAS. In particular, we need to ensure that only \textit{regular} dilation factors are generated, i.e., that the time-steps gaps between consecutive convolution inputs are all equal for a given layer. For example, we do not want to obtain a layer that takes as input time-steps $t$, $t-1$, $t-3$, and $t-10$, corresponding to gaps of 0, 1, and 6 time-steps respectively. In fact, such a layer would not be supported by most inference libraries, in particular those for edge devices~\cite{CubeAI,NNTool}, which only implement regular dilation, as the latter enables more regular memory access patterns and better low-level optimizations.

Based on these observations, we follow an approach similar to the one described in  Section~\ref{subsec:rf_search}.
We start from an array of trainable parameters $\gamma$, %
which are then combined to compose differentiable binary masks\footnote{In our preliminary work of~\cite{risso2021pit} we used a different mechanism to generate dilation masks as a \textit{product} of $\gamma$ elements instead of a sum. However, we found that this new approach is superior as it does not introduce nonlinear terms in the $\gamma$ gradients.}.
Our method only supports power-of-2 dilation factors which, besides being the most commonly used values, also simplify the generation of the masks.
Thus, we have: $len(\gamma) = \left \lceil \log_2(F_{seed})\right \rceil$.

In order to obtain the elements of $\Theta_{\Gamma}$, we pass through an intermediate array $\Gamma$, generated similarly to (\ref{eq:beta_aggregation}):
\begin{equation} \label{eq:gamma_aggregation}
    \Gamma_{i} = \mathcal{H} \left( \sum_{j = 1}^{len(\gamma) - i} |\gamma_{len(\gamma) - j}| \right)
\end{equation}
Then, the mask is obtained  by further reorganizing the $\Gamma_{i}$ values into the vector $\Theta_\Gamma$, of length $F_{seed}$, as follows:
\begin{equation}
\Theta_{\Gamma,i} = \Gamma_{k(i)},\text{ with } k(i) = \sum_{p=1}^{len(\gamma)} 1 - \delta(i\textrm{ mod } 2^p, 0)
\end{equation}
and where $\delta()$ is Kronecker's Delta function.
% \begin{equation}
% \Theta_{\Gamma,i} = \Gamma_{k(i)},\text{ with } k(i) = \sum_{p=1}^{len(\gamma)} \text{NotMultiple}(i, 2^p)
% \end{equation}
% %
% and where ``NotMultiple'' is a function that returns:
% \begin{equation}
% \text{NotMultiple}(a, b) =
% \begin{cases}
% 1\text{ if } a \mod b \ne 0\\
% 0\text{ otherwise }\\
% \end {cases}
% \end{equation}
%
This reorganization ensures that the $\Gamma$ element with the largest index ($\Gamma_{len(\gamma) - 1}$) ends up in all positions corresponding to time-steps that would be skipped by a layer with $d=2$. Similarly, the element with the second largest index ends-up in positions that are skipped when using $d=4$, and so on. This, combined with the fact that, by construction of (\ref{eq:gamma_aggregation}), it holds that $\Gamma_i \le \Gamma_j$ for $i > j$, ensures that the dilation is \textit{progressively increased}. In other words, each new $\Gamma_i$ binarized to 0 increases the dilation by a factor 2.

\begin{figure}[t]
  \centering
  \includegraphics[width=1.\columnwidth]{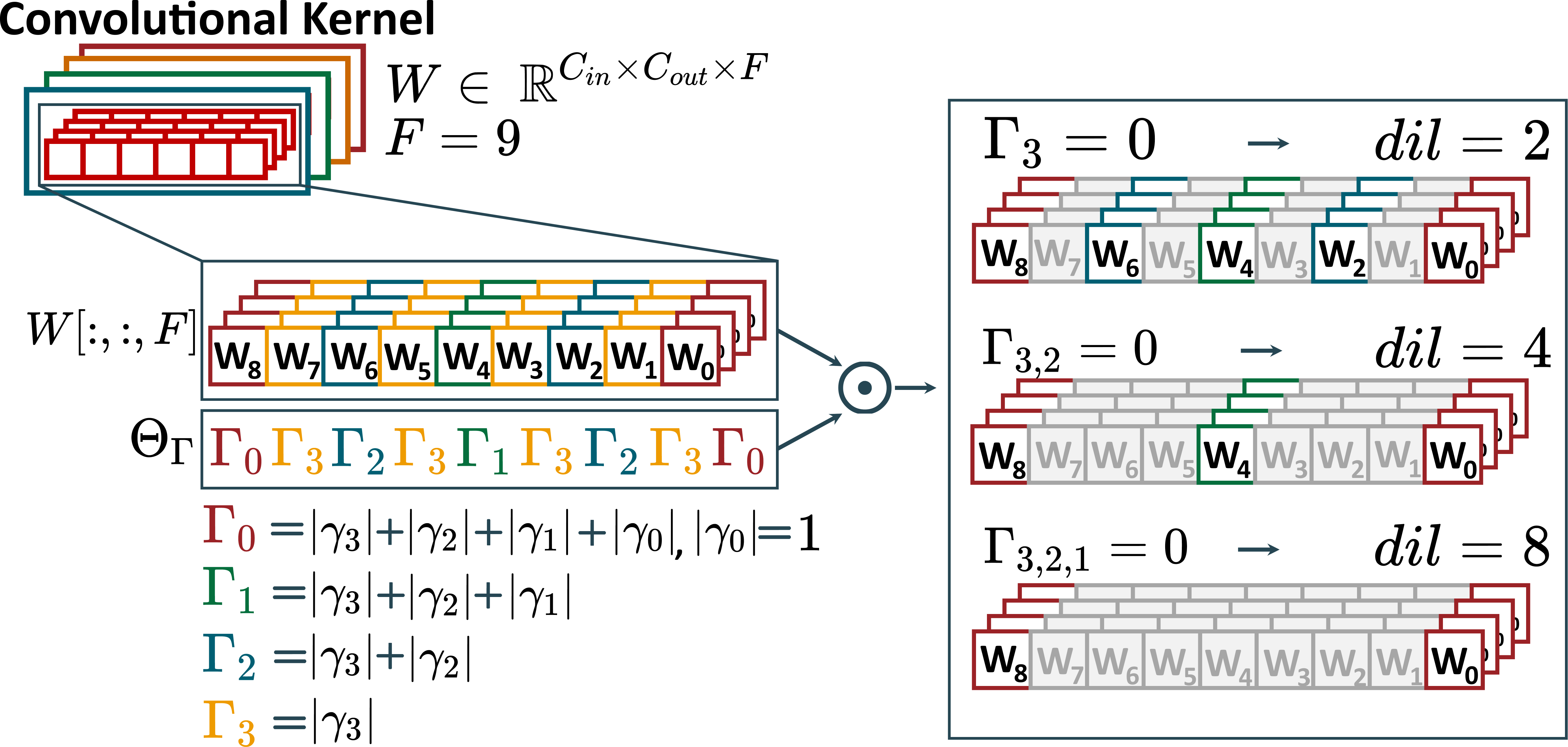}
  %\vspace{-0.4cm}
  \caption{Dilation search example. Each $\Gamma_i$ = 0 increases $d$ by a factor 2.}
  \label{fig:dil_search}
\end{figure}

The obtained $\Theta_\Gamma$ vector is multiplied with the $W$ tensor, exactly as in (\ref{eq:beta_mask}), after setting the seed layer dilation to 1.
Again, we fix $\gamma_0=1$ to ensure that we never prune the entire convolution.
An example of how the tensor is generated and of its effect on the dilation is shown in Figure~\ref{fig:dil_search}. 
%
% \begin{figure}[ht]
%   \centering
%   \includegraphics[width=1.\columnwidth]{figures/dil_mask.pdf}
%   %\vspace{-0.4cm}
%   \caption{.}
%   \label{fig:dil_mask}
% \end{figure}
% %
%
In practice, similarly to the receptive field mask, also $\Theta_{\Gamma}$ is obtained from $\gamma$ with a simple matrix multiplication:
\begin{equation} \label{eq:gamma_mask}
    \Theta_{\Gamma} = \mathcal{H} (C_{\gamma} \cdot | \gamma |)
\end{equation}
where $C_{\gamma}$ is a constant matrix composed of 0s and 1s that can be generated procedurally based on the value of $F_{seed}$. An example of $C_\gamma$ is shown on the right of Figure~\ref{fig:matrix_transf}.

\subsubsection{Joint Search}\label{subsec:joint_search}

In order to jointly optimize all three aforementioned hyper-parameters, we simply apply all three $\Theta$ masks to the weight tensor of a layer. Therefore, the equivalent of equation (\ref{eq:1d_conv}) for a seed convolutional layer during a joint search is:
\begin{equation}
    y_t^m = \sum_{i=0}^{K-1} \sum_{l=0}^{C_{in}-1} x_{ts-i}^l \cdot (\Theta_{B,i} \odot \Theta_{\Gamma,i} \odot (\Theta_{A,m} \cdot W_i^{l,m}))
\end{equation}
Note that, as anticipated in Section~\ref{subsec:dil_search}, we set the seed layer dilation to 1, since we want to let PIT explore all possible $d$ values. In our experiments, we found that performing such a joint search yields superior results with respect to optimizing the three hyper-parameters sequentially, since PIT can take into account the complex interactions among them (especially among $F$ and $d$), see Section~\ref{subsec:abl_study}.

\subsection{Regularization} \label{subsec:regularizer}
Following the same approach of state-of-the-art DNASes~\cite{proxylessnas_2018,fbnetv2_2020,morphnet_2018} PIT searches for accurate yet low-complexity architectures by combining the task-specific loss function $\mathcal{L}$ with a regularization term $\mathcal{R}$ as in (\ref{eq:dnas}).
The additional differentiable term encodes a prior in the loss landscape that directs the optimization towards low-cost solutions.
%
% that minimize, at the same time, both $\mathcal{L}$, related to performance, as well as $\mathcal{R}$.
%
% In our NAS $\mathcal{R}$ is a term which is proportional to a certain \textit{cost metric} that we want to optimize.
%
The two cost metrics considered in this work are the number of parameters (or size) of the model, and the number of operations (OPs) for an inference.
The corresponding two regularizers $\mathcal{R}_{\mathit{size}}$ and $\mathcal{R}_{\mathit{ops}}$ are differentiable functions of the \textit{pre-binarization} masks $\Tilde{\Theta}_A$, $\Tilde{\Theta}_B$ and $\Tilde{\Theta}_\Gamma$, i.e., the outputs of (\ref{eq:alpha_mask}), (\ref{eq:beta_mask}) and (\ref{eq:gamma_mask}) but without the Heaviside binarization. The latter, in turn, depend on the trainable architectural parameters $\alpha$, $\beta$ and $\gamma$. 
We use pre-binarization masks as in~\cite{morphnet_2018}, because this yields a smoother loss landscape, improving convergence. The details of the two regularizers are provided below.
%
% The same idea is pursued in our work, having that the regularization terms are function of $\alpha$, $\beta$ and $\gamma$: $\mathcal{R} = \mathcal{R}(\alpha, \beta, \gamma)$.
%
% Moreover, each NAS parameters $\theta_{i}$ is treated with a Lasso regularization term, to promote their sparsification enabling the exploration of smaller sub-architectures contained in the seed.
%
\subsubsection{Size Regularizer}
The Size Regularizer $\mathcal{R}_{\mathit{size}}$ estimates, during each forward-pass, the \textit{effective} number of parameters of the network, based on the values of the differentiable binary masks.
The number of parameters of a convolutional layer, i.e., the size of weight tensor $W$, is equal to $C_{in} \times C_{out} \times K$.
Accordingly, we define the size regularizer for a TCN with $N$ convolutional (or FC) layers as:
\begin{equation} \label{eq:size_reg}
    \mathcal{R}_{\mathit{size}} = \sum_{n=0}^{\text{N-1}} (\mathcal{R}_{\mathit{size}}^{(n)}) = \sum_{n=0}^{\text{N-1}} C_{out,eff}^{(n-1)} \cdot C_{out,eff}^{(n)} \cdot K_{eff}^{(n)}
\end{equation}
where:
\begin{equation}\label{eq:cout_eff}
C_{out,eff}^{(n)}  = \sum_{i=0}^{C_{out,seed}^{(n)}-1} \Tilde{\Theta}_{A,i}^{(n)}
\end{equation}
is the effective number of channels in the $n$-th layer, and:
\begin{equation}\label{eq:k_eff}
K_{eff}^{(n)} =  \sum_{i=0}^{F_{seed}^{(n)}-1}\frac{\Tilde{\Theta}_{B,i}^{(n)}}{F_{seed} - i} \cdot \frac{\Tilde{\Theta}_{\Gamma,i}^{(n)}}{len(\gamma) - k(i)}
\end{equation}
is the effective kernel size, which depends both on the total receptive field and on the dilation.
For the 1st layer of the network, $C_{out,eff}^{(n-1)}$ is constant and equal to the number of channels of the input signal.

The definitions of (\ref{eq:cout_eff}) and (\ref{eq:k_eff}) are continuous relaxations of the number of \textit{active} (non-pruned) channels and time-slices of $W^{(n)}$ respectively.
By minimizing $R_{size}$, PIT is encouraged to reduce the $\Tilde{\Theta}$ values, bringing them below the binarization threshold. Depending on the regularization strength $\lambda$ of (\ref{eq:dnas}) PIT balances the corresponding reduction in cost with the accuracy drop caused by eliminating $W$ slices from the layer, reducing only the $\Tilde{\Theta}$ elements associated to unimportant slices.

The denominators in (\ref{eq:k_eff}) are needed to make sure that, when $\beta$ and $\gamma$ are equal to 1 (i.e., the initialization value, see Section~\ref{subsec:train_proc}), $K_{eff}^{(n)}$ corresponds to the real filter size of the seed. In fact, each $\Tilde{\Theta}_{B/\Gamma}$ is obtained as sum of a different number of $\gamma$ (or $\beta$) elements. As a result, without normalization, the estimated cost would be higher than the real filter size.
For instance, in a layer with $F_{seed} = 5$ and with all $\beta$/$\gamma$ initialized at 1, without the denominators, we would have $K_{eff}$ = 33, which is clearly incorrect. Conversely, with the denominators, we have $K_{eff} = 5 = F_{seed}$, which is correct, since the initialization of $\gamma=1$ implicitly imposes $d=1$.

\subsubsection{OPs Regularizer}
The second proposed regularizer $R_{ops}$ estimates the number of operations required to perform an inference.
Since the number of OPs of a 1D convolutional layer is $T \times C_{in} \times C_{out} \times K$, where $T$ is the output sequence length defined in (\ref{eq:1d_conv}), the regularizer expression is simply:

\begin{equation} \label{eq:flops_reg}
\begin{aligned}
    \mathcal{R}_{\mathit{ops}} = \sum_{n=1}^{\text{N}}(\mathcal{R}_{\mathit{size}}^{(n)} \cdot T^{(n)})
\end{aligned}
\end{equation}
In practice, when targeting the reduction of the total OPs for inference, the only difference in the regularizer is that the cost of each layer is weighted by the output sequence length. This is particularly important in presence of layers such as pooling, strided convolution, etc., which significantly reduce $T$, and consequently the number of OPs for the downstream part of the network.

\subsection{Training Procedure} \label{subsec:train_proc}
\input{algorithms/all_in_one_alg}
Algorithm~\ref{alg:nas_search} summarizes the three main phases of a PIT architecture search.
The first phase consists of $\text{Steps}_{\text{wu}}$ iterations of warmup.
At this stage of the algorithm, all $\theta$ parameters (i.e., $\alpha$, $\beta$ and $\gamma$) are initialized to 1 and frozen. Accordingly, all elements of the binary masks $\Theta$ are also binarized to 1.
Therefore, warmup coincides with a normal training of the seed network, where the only objective is minimizing the task loss function $\mathcal{L}$. 
The number of warmup iterations is a user-defined parameter.
%
% that can vary between 0 (i.e., no warmup) and \textit{max}, where \textit{max} denotes a training of the seed network until convergence.
%
In practice, in all our experiments, we warm up to convergence.

The second phase is where the actual NAS takes place. In the search loop,  the model weights $W$ and the architectural parameters $\theta$ are optimized simultaneously.
Accordingly, the goal of this phase is to minimize the sum of the task-specific loss $\mathcal{L}$ and of one of the two the regularization losses $\mathcal{R}$ discussed in Section~\ref{subsec:regularizer}, weighted by the regularization strength $\lambda$.
The duration of the search phase is controlled by an early-stop mechanism which monitors the value of $\mathcal{L}$ on an unseen validation split of the target dataset, and stops the search when the latter does not improve for 20 epochs.

Finally, in the third and last phase the $\theta$ parameters, and corresponding $\Theta$ binary masks, are frozen to their latest values.
This corresponds to sampling from the search space the architecture that PIT determined as optimal during the previous phase.
Then, the weights $W$ of the selected network are fine-tuned or re-trained from scratch, considering only the $\mathcal{L}$ loss.

In order to obtain different Pareto points in the accuracy versus cost (size or OPs) space with PIT, it suffices to repeat Algorithm~\ref{alg:nas_search} changing the regularization strength $\lambda$. More precisely, the warmup phase can be performed just once, saving the final weights of the seed network. Overall, Algorithm~\ref{alg:nas_search} has a complexity that is comparable to a single TCN training. Moreover, the requirements in terms of GPU time and memory are greatly reduced with respect to a supernet-based DNAS. Therefore, obtaining 10s of Pareto points by changing $\lambda$ still has a manageable cost, as shown for example by the results of Figure~\ref{fig:timing_comparison}.

% since, except for the negligible overhead due to the additional architectural parameters $\theta$, the seed is not larger than the most complex TCN that PIT can produce as output ( $\lambda=0$).

% Starting from the receptive field of the  is fully defined specifying two additional parameters:
% \begin{itemize}
%     %
%     \item the \textit{expansion rate}, $e$, is an integer which is multiplied to the value of $F^s$ to arbitrarily enlarge the search space.
%     %
%     The effective starting receptive field size used in the seed network will be $F = e \cdot F^s$.
%     %
%     \item the \textit{granularity}, denoted as $\mathit{gr}$, is an integer $1 \le \mathit{gr} \le (F - 1)$, which defines how \textit{fine-grained} is the search performed.
%     %
% \end{itemize}
% %
% The search is performed starting from a receptive field with $F = e \times F^s$, $d=1$ and $F \equiv K$.

%% file: tables/notation_recap.tex
\begin{table}[t]
\centering
\caption{List of symbols used in the paper.}\label{tab:notation}
\begin{adjustbox}{max width=1\columnwidth}
\begin{tabular}{|>{\raggedright\arraybackslash}p{0.15\columnwidth}|p{0.8\columnwidth}|}
\hline
\multicolumn{1}{|c|}{\textbf{Symbol}} & \multicolumn{1}{c|}{\textbf{Description}} \\ \hline
$x$ & Input activations of a convolutional layer\\
$y$ & Output activations of a convolutional layer\\
$T$  & Output sequence length of a convolutional layer\\
$C_{in}$, $C_{out}$ & Number of input/output channels of a conv. layer\\
$W$ & Convolutional filter weights \\
$K$ & Convolution filter size\\
$s$ & Convolution stride\\
$d$ & Convolution dilation\\
$F$ & Convolution receptive field \\
$\mathcal{L}$ & Task-specific loss function\\
$\mathcal{R}$ & Regularization loss function \\
$\lambda$ & Regularization strength \\
$\mathcal{S}$, $\hat{\mathcal{S}}$ & Search space and sampled architecture \\
$L_{n}$ & Generic convolutional/FC layer\\
$N$ & Number of convolutional/FC layers \\
% $\mathcal{H}$ / $\mathit{th}$ & Heaviside step function and its fixed threshold \\
$\theta$, $\Theta$ & Generic NAS architectural parameters and corresponding binary mask \\
$\alpha$, $\Theta_{A}$ & NAS architectural parameters to optimize $C_{out}$ and corresponding binary mask \\
$\beta$, $\Theta_{B}$ & NAS architectural parameters for $F$, and corresponding binary mask \\
$\gamma$, $\Gamma$, $\Theta_{\Gamma}$ & NAS architectural parameters for $d$, intermediate binary mask elements and final binary mask \\
% $\Tilde{\Theta}_A$ / $\Tilde{\Theta}_B$ / $\Tilde{\Theta}_\Gamma$ & Pre-binariaztion masks \\
$C_{\beta}$, $C_{\gamma}$ & Transformation matrices to generate $\Theta_{B}$ and $\Theta_{\Gamma}$ from $\beta$ and $\gamma$. \\
$k(i)$ & Index mapping function used to generate $\Theta_{\Gamma}$ from $\Gamma$ \\\hline
% $\delta()$ & Kronecker's Delta function \\\hline
\end{tabular}
\end{adjustbox}
\end{table}

%% file: algorithms/all_in_one_alg.tex
\begin{algorithm}[t]
\begin{algorithmic}[1]
\caption{\label{alg:nas_search}}
    \For{$i \gets 1, \dots, \rm Steps_{wu}$} {\color{darkspringgreen}\#warmup loop}
        \State Update $W$ based on $\nabla_{W} \mathcal{L}(W)$
    \EndFor
    \While{not converged} {\color{darkspringgreen}\#search loop}
        \State Update $W$ and $\theta$ based on $\nabla_{W, \theta} (\mathcal{L}(W; \theta) + \lambda \mathcal{R}(\theta))$
    \EndWhile
    \For{$i \gets 1, \dots, \rm Steps_{ft}$} {\color{darkspringgreen}\#fine-tuning loop}
        \State Update $W$ based on $\nabla_{W} \mathcal{L}(W)$
    \EndFor
\end{algorithmic}
\end{algorithm}

%% file: sections/04_Benchmark.tex
\revFloatRemove{
% \begin{figure*}[t]
%   \centering
%   \includegraphics[width=.7\textwidth]{figures/seeds.pdf}
%   \caption{\revRemove{Seed network architectures for the four considered benchmarks.}}
%   \label{fig:seeds}
% \end{figure*}
% %
%
\begin{figure*}[t]
  \centering
  \includegraphics[width=.8\textwidth]{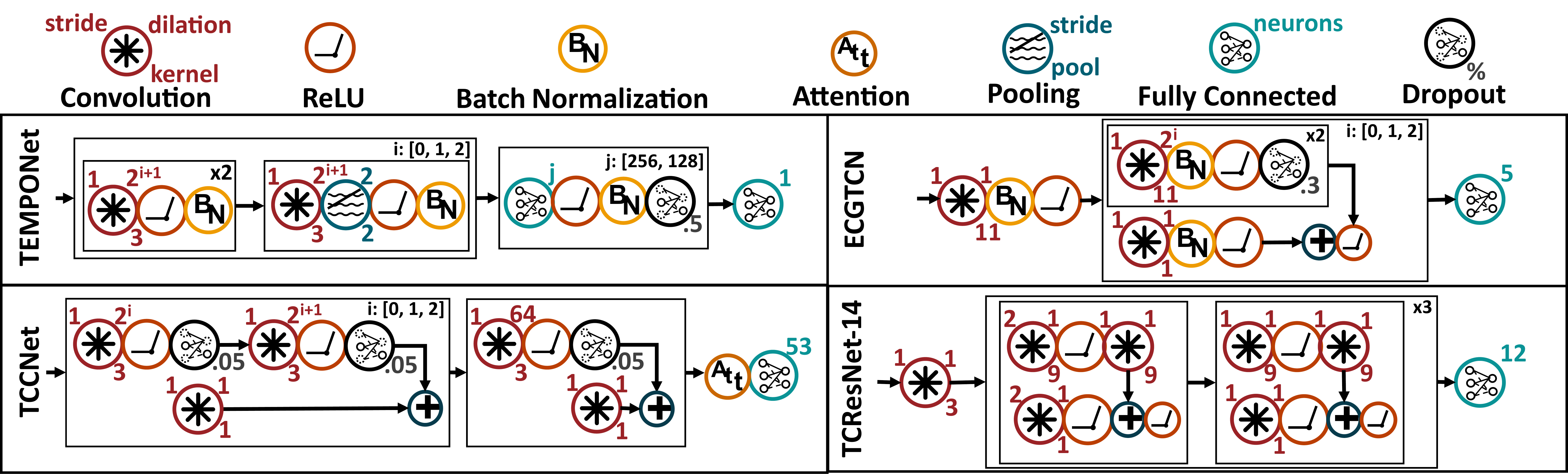}
  \caption{\revRemove{Seed network architectures for the four considered benchmarks.}}
  \label{fig:seeds}
\end{figure*}
}

We test PIT on four edge-relevant real-world benchmarks. We select diverse benchmarks to comprehensively evaluate the effectiveness of the proposed NAS. Specifically, we consider regression as well as classification tasks; \rev{the inputs analyzed are both raw or extracted features}, and the TCNs employed as seed are based on different architectural styles.
\revRemove{The four benchmarks are described in detail in the rest of this section.}

\subsection{PPG-based Heart-Rate Monitoring} \label{subsec: dalia_bench}
The first benchmark \revRemove{considered }deals with Heart-Rate (HR) monitoring on wrist-worn devices, using Photoplethysmography (PPG) sensors coupled with tri-axial accelerometers to mitigate the effect of motion artifacts~\cite{reiss2019deep,risso2021robust}.
%
% While this type of HR tracking is very common in commercial devices, its accuracy is strongly impaired by Motion Artifacts (MAs), i.e., PPG signal distortions caused by wrist movements.
% %
% To mitigate this problem, state-of-the-art approaches couple the PPG signal with tri-axial acceleration data, feeding them both to a DNN that performs a regression to obtain a HR estimation~\cite{reiss2019deep,risso2021robust}.
%
% In our experiments,
%
% \revRemove{We target the PPG-Dalia~\cite{reiss2019deep} dataset, 
% which includes PPG signals and 3D acceleration recordings collected during daily life activities on a population of 15 subjects. The}
%
\rev{We target the PPG-Dalia~\cite{reiss2019deep} dataset, and the}
task is formulated as a \textit{regression} of the HR value, whose ground truth is derived with ECG measurements.
%
% as well as the corresponding ground-truth HR (derived from ECG measurements).
%
%as well as ECG traces used to compute the ground-truth HR.
%
% The data are collected during daily life activities on a population of 15 subjects.
%
\revRemove{Both PPG and acceleration are sampled at \unit[32]{Hz} and organized in \unit[8]{s} long sliding windows with a time shift of \unit[2]{s} between successive windows.} \rev{All results refer to the same input windowing and cross-validation scheme proposed in~\cite{reiss2019deep}.}

The seed network for this task is TEMPONet, a TCN originally proposed in \cite{temponet_2019} and later used for HR monitoring with state-of-the-art results in~\cite{risso2021robust}.
\revRemove{The network architecture is depicted in the top left of Figure~\ref{fig:seeds}.}
\rev{The network is composed of three \textit{feature extraction} blocks and a final \textit{regressor} module with three FC layers.}
Each feature extraction block is\revRemove{ in turn} made of three convolutional layers with BatchNorm and ReLU activation, followed by an average pooling.
The \revRemove{regressor }FC layers are also followed by \revRemove{batch-normalization} \rev{BatchNorm} and ReLU, and by a dropout layer with \unit[50]{\%} rate. 
With respect to the original TEMPONet, our seed \revRemove{TCN }is obtained doubling the receptive field of all \revRemove{convolutional layers }\rev{convolutions} and setting the dilation \revRemove{factor }to 1.
%
% , to give PIT maximum freedom in optimizing both these parameters.

% This gives PIT the freedom to also search also for networks \textit{larger} than the original TEMPONet, with longer filters or smaller dilation.

%
\subsection{ECG-based Arrhythmia Detection} \label{subsec: ecg_bench}
Our second benchmark deals with Electrocardiogram (ECG)-based arrhythmia detection, for wearable medical devices.
%
% Usually the analysis of ECG signals is performed not in real-time.
% %
% Performing it directly in-place, with high accuracy, on wearable devices would be extremely helpful for personal health-care monitoring enabling a faster diagnosis of severe heart diseases. 
%
% \revRemove{We target the ECG5000 dataset~\cite{chen2015general}, which includes 5000 ECG records subdivided in 500 points used as training set and 4500 points used as test set.
% %
% The task consists in classifying properly the ECG signals choosing between 5 different labels: Normal (N), R-on-T Premature Ventricular Contraction (Ron-T PVC), Premature Ventricular Contraction (PVC), Supraventricular Premature or Ectopic beat (SP or EB), and unclassified beat (UB).}
%
\rev{We target the ECG5000 dataset~\cite{chen2015general}, and the task consists in classifying the ECG signals in 5 classes: Normal, R-on-T Premature Ventricular Contraction, Premature Ventricular Contraction, Supraventricular Premature or Ectopic beat, and Unclassified Beat.}

The reference TCN \revRemove{for this task }is ECGTCN, originally proposed in~\cite{ingolfsson2021ecg}\revRemove{, and shown in the top right of Figure~\ref{fig:seeds}}.
Differently from TEMPONet, ECGTCN\revRemove{'s architecture} is based on residual blocks.
\rev{It has a first convolutional layer that enlarges the number of input channels, followed}
by three modular blocks, each including two dilated convolutions with ReLU activation, \rev{BatchNorm and 50\% dropout}.
The input and output feature maps of each block are then summed together.
%
% to form the residual connection.
%
When the number of input and output channels differs, the residual path also includes a point-wise convolution (i.e., $K = 1$) in order to adapt the tensor sizes.
The PIT seed is obtained from ECGTCN,
%
% keeping the same number of output channels and the original receptive field, while
%
%
setting the dilation \revRemove{factor }of all layers to 1, while keeping the original receptive field.

\subsection{sEMG-based Hand-Gesture Recognition} \label{subsec: ninapro_bench}
The third benchmark deals with hand-gesture recognition based on  surface electromiography (sEMG) signals.
\revRemove{The execution of such task at the edge is a key enabler for applications such as complex human-computer interfaces, non-invasive prosthesis control and rehabilitation.}
For this task, we target the NinaPro DB1 dataset~\cite{atzori2012building},
\revRemove{which includes records of 27 healthy patients monitored with 10 electrodes, while performing 52 heterogeneous hand-gestures, including basic finger and wrist movements, different hand poses, and grasping.
We use}
\rev{which includes 52 heterogeneous gesture classes, using} the same data pre-processing and augmentation \revRemove{scheme }described in~\cite{tsinganos2019improved}.

The seed network is TCCNet, originally proposed in~\cite{tsinganos2019improved}\revRemove{, and depicted in the bottom left of Figure~\ref{fig:seeds}}.
The architecture includes three feature extraction blocks, each composed of two dilated convolutions with ReLU and dropout (5\% rate\revRemove{ in this case}) and a residual branch with a point-wise convolution.
The classifier includes an attention layer of the type described in~\cite{yang2016hierarchical} and a final\revRemove{ fully-connected} \rev{FC} layer with 53 output neurons (52 hand-gestures + 1 unknown class).
\revRemove{As before, t}\rev{T}he PIT seed is obtained simply setting the dilation to 1 in all layers.

\subsection{Keyword Spotting} \label{subsec: gspeech_bench}
Our last benchmark is keyword spotting (KWS)\revRemove{, a key component of speech-based human-machine interfaces (e.g., for smart personal assistants)}.
%
% Notable KWS applications are represented by systems including a wake-word detection mechanism as that used in smart personal assistant (e.g., "Okay Google"\texttrademark, "Hey Siri"\texttrademark, "Alexa"\texttrademark).
%
%
We target the \revRemove{standard benchmark for KWS systems, i.e., the }Speech Commands v2 dataset~\cite{warden2018speech}, \revRemove{which consists of 105829 utterances collected from 2618 speakers.
%
% with different accents.
%
%, including 30 different words and background noise.
%
We follow}
\rev{following}
the pre-processing scheme proposed by the MLPerf Tiny 
\revRemove{industry-standard }benchmark suite~\cite{banbury2020benchmarking} which produces 12 possible labels, including 10 words and two special classes for "unknown" and "silence".

As seed, we use the TCN presented in~\cite{choi2019temporal}, called TC-ResNet14\revRemove{, whose architecture is shown in the bottom right of Figure~\ref{fig:seeds}}.
%
% The proposed architecture present a similar structure to other TCNs with residual connections. 
%
The main difference with the other reference TCNs is that the original TC-ResNet14 did not use dilation, and the modular convolutional blocks alternate plain convolutions with strided convolution with $s = 2$. PIT's seed is obtained doubling the receptive field in each layer.

%% file: sections/05_Results.tex
\begin{figure*}[ht]
  \centering
  \includegraphics[width=.9\textwidth]{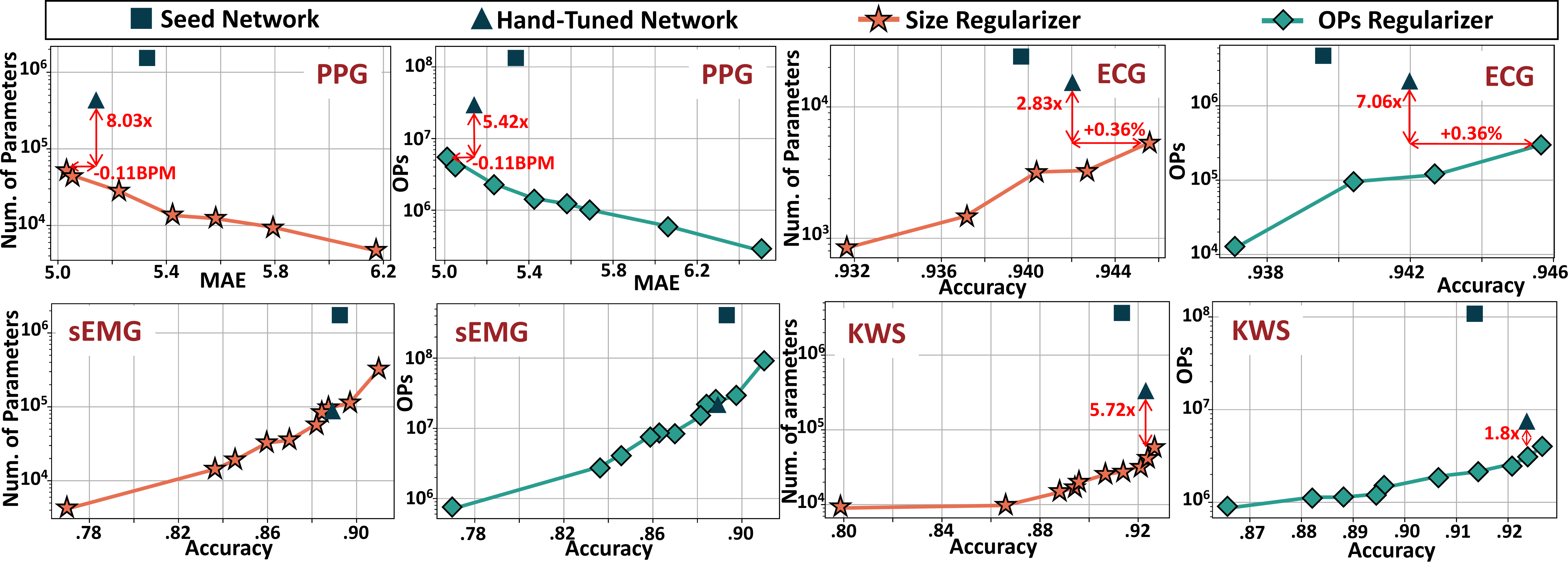}
  %\vspace{-0.4cm}
  \caption{\rev{Overall PIT Pareto fronts for the four target benchmarks, and comparison with seed and hand-tuned TCNs.}}
  \label{fig:pareto_fronts}
\end{figure*}
This section discusses the results obtained by PIT on the four aforementioned benchmarks.
%
% The exploration of each search space has been conducted using different regularization objectives (i.e., size and FLOPs) in order to find good candidate networks for an embedded deployment.
%
In particular, in Section~\ref{subsec:search_space_expl}, we present the global results of our NAS search in the accuracy versus number of parameters and accuracy versus number of OPs planes.
\rev{In Section~\ref{subsec:abl_study} we conduct ablation studies on one of the benchmarks, and
in Section~\ref{subsec:sota_comp} we compare our approach with a state-of-the-art DNAS, ProxylessNAS \cite{proxylessnas_2018}, and with two state-of-the-art DMaskingNAS approaches, namely, MorphNet~\cite{morphnet_2018} and FBNetV2~\cite{fbnetv2_2020}. Since the code for~\cite{fbnetv2_2020} is not publicly available, we re-implemented it based on the information provided in the paper.}
Finally, Section~\ref{subsec:deployment} presents the  memory, latency and energy consumption results obtained deploying some of the networks found by PIT on two commercial edge devices.

PIT is written in Python (v3.6) and it is based on PyTorch (v1.7.1). All our training experiments and NAS searches are performed on a single NVIDIA TITAN Xp GPU with 12GB memory.
The two deployment targets considered are: i) the multicore GAP-8 IoT processor by GreenWaves Technologies~\cite{flamand2018gap} and ii) the single-core STM32H7 MCU by STMicroelectronics\cite{stm32h7}. 
As inference software backend, we use the open-source layers library of~\cite{Burrello2021} coupled with the tiling tool of~\cite{burrello2021dory} for GAP-8, and the CMSIS-NN library~\cite{cmsis-nn}
%
%(modified to support dilation on 1D convolutions)
%
for the STM32H7. All deployed networks are quantized to 8-bit, using PyTorch's built-in quantization algorithm.

\subsection{Search Space Exploration} \label{subsec:search_space_expl}

% In this section we propose and discuss the results obtained applying our NAS on the aforementioned tasks.
% %
% The search space defined by each seed network has been explored following the scheme detailed in Alg.~\ref{alg:nas_search} with a number of warmup epochs such that the network is trained until convergence (i.e., performance not improving on a validation split for a pre-defined number of epochs). 
%
Figure~\ref{fig:pareto_fronts} shows the results of applying PIT to the four benchmarks. The graphs report the TCNs accuracy (for classification tasks) or Mean Absolute Error (MAE, for regression tasks) on the x axis, and the number of parameters or OPs per inference on the y axis.
The curves correspond to the outputs of PIT, where different points are obtained varying the regularization strength $\lambda$ and considering both size and OPs regularizers.
%
% The results are then merged to form Pareto fronts.
%
Moreover, each plot also reports the metrics of two additional TCNs.
Black triangles correspond to the results obtained by the \textit{hand-tuned} state-of-the-art TCNs \revRemove{of Figure~\ref{fig:seeds},} directly taken from \cite{risso2021robust, ingolfsson2021ecg, tsinganos2019improved, choi2019temporal}, with the original number of channels, receptive fields, and dilation factors. Black squares, instead, indicate the metrics of the PIT \textit{seeds}, i.e., the same networks modified as described in Section~\ref{sec:benchmark} (setting $d=1$ everywhere, etc.) to enlarge the PIT search space.

The upper-left part of Figure~\ref{fig:pareto_fronts} reports the results on the PPG-DaLia dataset for the PPG-based HR monitoring task. This is the only regression task considered, so the network performance is measured with the MAE, for which lower values are better. As shown by the graphs, starting from a single seed network, PIT is able to obtain a rich collection of Pareto-optimal architectures, spanning more than one order of magnitude both in terms of parameters (4.7k-78k) and OPs (0.27M-9.6M). Notably, PIT networks dominate in the Pareto sense both the seed architecture and the hand-tuned state-of-the-art TEMPONet. In particular, we obtain a similar MAE to the seed TCN (5.38 vs 5.40 BPM), with 120.0$\times$ less parameters and 96.0$\times$ less operations. \rev{Moreover, PIT also finds a new state-of-the-art deep learning model for this task, achieving a MAE of just \unit[5.03]{BPM} while requiring only 53k parameters and 5.1M OPs, improving the best performing architecture proposed in~\cite{risso2021robust}\footnote{Note that this result is achieved without applying any additional post-processing as described in \cite{risso2021robust}.} requiring $8.03\times$ and $5.42\times$ less parameters and OPs.}

The Upper-right pair of charts shows the results obtained on the ECG5000 dataset for Arrhythmia Detection.
%
% Differently from the previous case the main metric considered here to asses the performance of the models is accuracy, then we will consider higher results as better.
%
%
PIT results span almost one order of magnitude in parameters (0.91k-5.36k) and OPs (50.3k-293.5k). Moreover, both the seed network and the hand-tuned one are Pareto-dominated.
The best performing architecture found by our NAS improves the accuracy of the hand-tuned network (+1.03\%) reducing both the number of parameters (-64.7\%) and the FLOPs (-85.8\%).

Lower-left part of Figure~\ref{fig:pareto_fronts} shows the results obtained for the sEMG-based Hand-Gesture Recognition task on the NinaPro-DB1 dataset. The richness and diversity of the found architectures in terms of size and number of OPs are similar to the previous two benchmarks. However, while PIT results still dominate the seed, in this case the hand-tuned TCNNet sits on the Pareto front.
Indeed, the PIT network that is nearest to the hand-tuned architecture on the curve, achieves a slightly lower accuracy (-0.47\%) traded-off with a reduction of size (-3.33\%). 
This result demonstrates the goodness of the original TCNNet proposed in~\cite{tsinganos2019improved} but, at the same time, it shows the good quality of the architectures found by our NAS, which despite starting from an oversized seed, is still able to produce optimized networks that closely resemble those tuned by experts.

Lastly, the lower-right part of Figure~\ref{fig:pareto_fronts} shows the two Pareto fronts obtained on the Google Speech Commands dataset for Keyword Spotting. Once again, we largely outperform both the seed and the hand-tuned TCNs. Specifically, the most accurate PIT architecture slightly improves the accuracy of the hand-tuned network (+0.36\%) while greatly reducing both the number of parameters (-82.53\%) and FLOPs (-44.53\%). Moreover, we obtain Pareto points that span 10k-98k parameters and 0.87M-3.98M OPs. It is important to note that the bad performance obtained by the seeds (black squares) for all four benchmarks is due to over-fitting, which in turn is caused by the large number of channels and receptive fields, and the absence of dilation.

\input{tables/table_lambda}

Table~\ref{tab:range_lambda} reports the range of regularizer strengths $\lambda$ used on the four benchmarks to obtain these results. In general, $\lambda$ should be set so that the two additive terms in the loss ($\mathcal{L}$ and $\lambda \mathcal{R}$) assume comparable values at the beginning of a training. This ensures that PIT takes into account both accuracy and inference cost in its search, without degenerating to one of the two corner cases, i.e., accuracy-driven-only and cost-driven-only optimization. The corresponding values of $\lambda$ vary for different tasks, as shown in the table. However, we found that a good rule of thumb, which works for all benchmarks, to identify the order of magnitude of the regularization strength is to start from $\lambda = $ 1/(Seed Model Size). Then, based on the results of a PIT search with this initial value, one can decide to increase/decrease $\lambda$ to obtain smaller/more accurate TCNs respectively. By monitoring the loss in the initial epochs, it is also very easy to detect when the NAS is falling in one of the corner cases (one term much larger than the other) and stop the search immediately, without wasting training time.

\subsection{Ablation Studies} \label{subsec:abl_study}

This section analyzes the impact of some of the most important PIT parameters. Due to space limitations, we report the results of this study only for the PPG-based HR monitoring benchmark.

\subsubsection{Hyper-parameters}

\begin{figure}[t]
  \centering
  \includegraphics[width=.8\columnwidth]{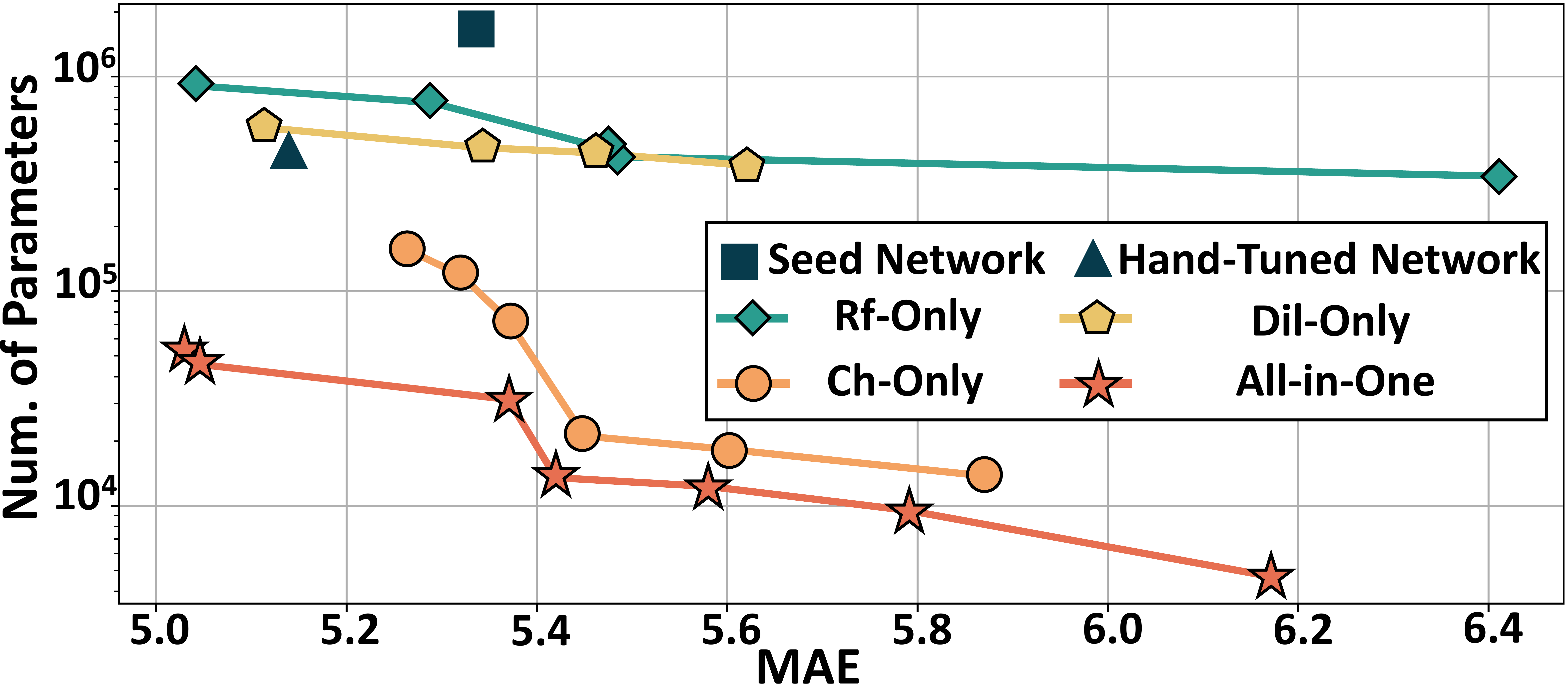}
  %\vspace{-0.4cm}
  \caption{\rev{Comparison between the results of PIT searches with different combinations of hyper-parameters for PPG-DaLia.}}
  \label{fig:dalia_ablation_search}
\end{figure}

Figure~\ref{fig:dalia_ablation_search} analyzes the contribution of different hyper-parameters to the quality of results found by PIT. \rev{For this experiment, we use the $\mathcal{R}_{size}$ regularizer and consider solutions in the MAE versus number of parameters space.}
We then repeat the NAS search 3 times. 
In each run, we freeze two of the three sets of architectural parameters ($\alpha$, $\beta$ and $\gamma$) to 1, letting PIT tune the third set.
This gives us: i) the results of a search that only optimizes the number of channels in each layer (\textit{Ch-Only}), performed on a TCN with maximal receptive field and $d=1$, ii) the results of a receptive field-only search (\textit{Rf-Only)}, on a TCN with maximal $C_{out}$ and $d=1$, and iii) the results of a dilation-only search (\textit{Dil-Only}) on a network with maximal $F$ and $C_{out}$.
The Pareto fronts obtained in each of these 3 conditions by varying the regularization strength $\lambda$ are shown in the figure, together with the output of a complete search that optimizes all three hyper-parameters simultaneously (\textit{All-in-One}). 

The results clearly show that the main source of parameters reduction and performance improvement is the search along the channels dimension.
This is probably due to the fact that the channels represent a large source of redundancy in hand-tuned TCNs,
%
%and CNNs in general,
%
since their number is typically set using common heuristics, irrespective of the target task (e.g., $C_{out}$ multiple of 32, progressively increasing along the depth of the network).
However, Figure~\ref{fig:dalia_ablation_search} also shows that optimizing \textit{only} the number of channels is not sufficient, and that a combined optimization that also consider receptive field and dilation can yield Pareto-optimal networks across the entire MAE/parameters range.

\subsubsection{Regularizers}
\begin{figure}[t]
  \centering
  \includegraphics[width=\columnwidth]{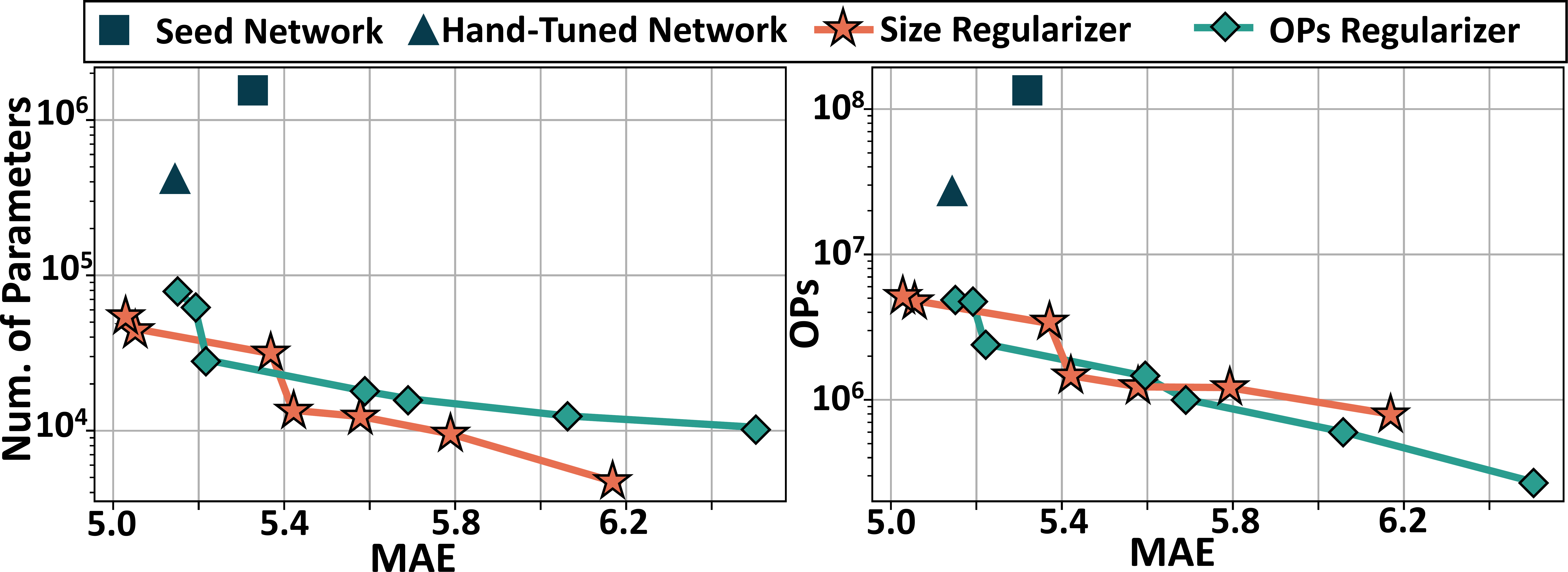}
  \caption{\rev{Comparison of $\mathcal{R}_{size}$ and $\mathcal{R}_{ops}$ regularizers for PPG-DaLia.}}
  \label{fig:dalia_ablation_flops_size}
\end{figure}
\rev{Figure~\ref{fig:dalia_ablation_flops_size} compares the Pareto fronts obtained using the $\mathcal{R}_{size}$ regularizer (with orange stars) and $\mathcal{R}_{ops}$ regularizer (with green diamonds).}
Note that the PPG-based HR monitoring benchmark is the one for which the distinction between model size and number of OPs is most relevant, due to the presence of several layers (average pooling and strided convolution) that modify the activation array length $T$.

The Figure shows that, as expected, the majority of the Pareto points in the MAE versus number of parameters plane are produced when using the $R_{size}$ regularizer, with the few exceptions being due to local minima. Vice versa, the $R_{ops}$ regularizer tends to generate superior solutions in terms of MAE versus number of OPs. 

\subsection{Comparison with state-of-the-art NAS tools} \label{subsec:sota_comp}
\begin{figure}[t]
  \centering
  \includegraphics[width=\columnwidth]{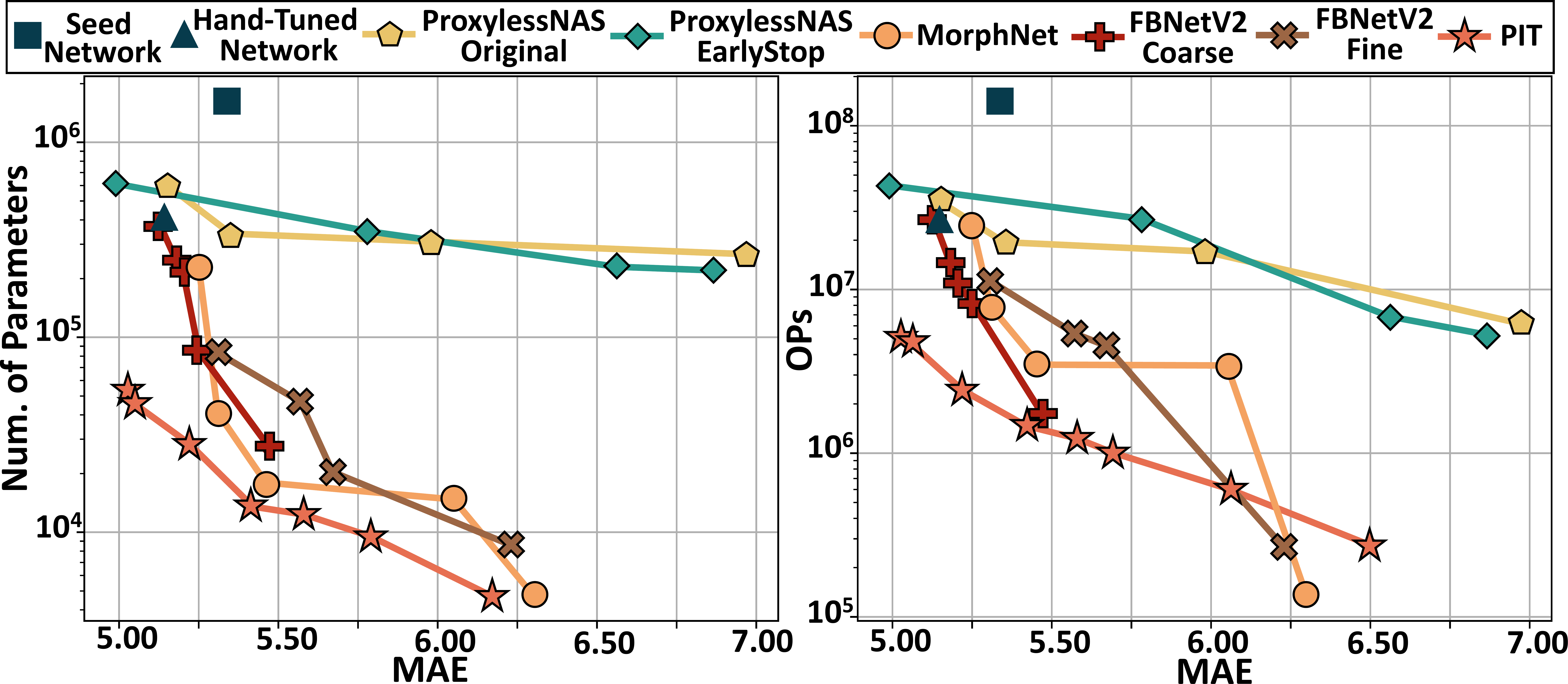}
  %\vspace{-0.4cm}
  \caption{\rev{Quality of results comparison between PIT and state-of-the-art NAS tools on the PPG-DaLia dataset.}}
  \label{fig:proxyless_perf_comp}
\end{figure}
\rev{Figure~\ref{fig:proxyless_perf_comp} compares the Pareto fronts obtained with PIT and three state-of-the-art NAS tools, namely ProxylessNAS \cite{proxylessnas_2018} MorphNet \cite{morphnet_2018} and FBNetV2~\cite{fbnetv2_2020}, on the HR monitoring benchmark.
Results show that PIT outperforms all three across the entire design space, except for one MorphNet and one FBNetV2 point, that achieve a very low number of operations, although at the cost of a quite large MAE.
The main reason for the superior results of PIT is the fact that our NAS explores a larger and finer-grain search space with respect to the baselines.
For what concerns MorphNet and FBNetV2, this is partly due to the intrinsic nature of those tools, which cannot explore receptive field, nor dilation~\cite{morphnet_2018,fbnetv2_2020}. Accordingly, $F$ and $d$ in their respective seeds have been set to the hand-tuned values of the state-of-the-art network. This different search starting point compared to PIT is the reason why, in the low-size/high-MAE regime, these tools find a single Pareto-optimal point.

For FBNetV2, we considered a \textit{Coarse} search space, including 4 $C_{out}$ alternatives per layer, uniformly spaced, i.e., $\nicefrac{1}{4}C_{out,seed}$, $\nicefrac{1}{2}C_{out,seed}$, $\nicefrac{3}{4}C_{out,seed}$ and $C_{out,seed}$.
and a \textit{Fine} search space, which instead evaluates all $C_{out}$ values with a granularity of 1. The latter is more similar to PIT, but the former achieves superior results in most cases. This is because FBNetV2 uses a pre-defined binary mask for each layer variant, combining them through a Gumbel softmax, as explained in Sec.~\ref{subsec:ch_search}. Experimentally, we found that with a too large number of masks, the search becomes unstable and yields sub-optimal results. In contrast, PIT does not have this limitation since it uses \textit{independent} trainable masks that keep or eliminate an individual channel.}

ProxylessNAS, \revRemove{instead, }being a super-net-based DNAS, would be virtually able to explore the entire PIT search space, as long as all the versions of layers to be explored are included in the super-net~\cite{proxylessnas_2018}. However, doing so would result in a too huge network, impossible to train due to memory and time requirements. In fact, as detailed in Section~\ref{sec:methods}, PIT explores $C_{out}$ and $F$ with a granularity of 1, and for $d$, it considers all possible power-of-2 values. Therefore, each super-net node should include $C_{out,seed} \cdot F_{seed} \cdot \lceil \log_2(F_{seed})\rceil$ different layers, connected in parallel. With the same parameters used for the example at the end of Section~\ref{subsec:search_space}, this would correspond to $\approx$10000 different versions of \textit{each layer}.

\begin{figure}[t]
  \centering
    \includegraphics[width=.85\columnwidth]{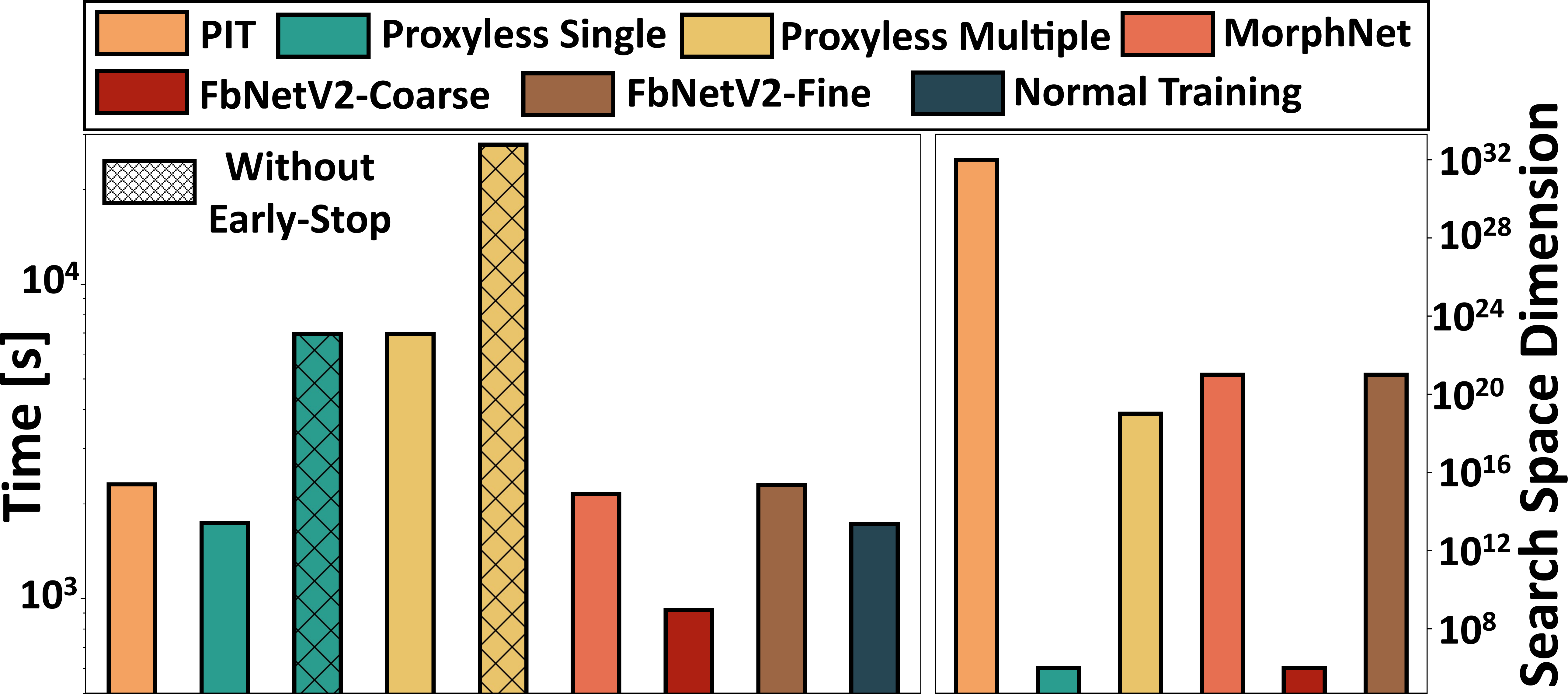}
  \caption{\rev{Search space and time comparison between PIT and state-of-the-art NAS tools on the PPG-DaLia dataset.}}
  \label{fig:timing_comparison}
  %\vspace{-0.5cm}
\end{figure}
\begin{figure*}[ht]
  \centering
  \includegraphics[width=.9\textwidth]{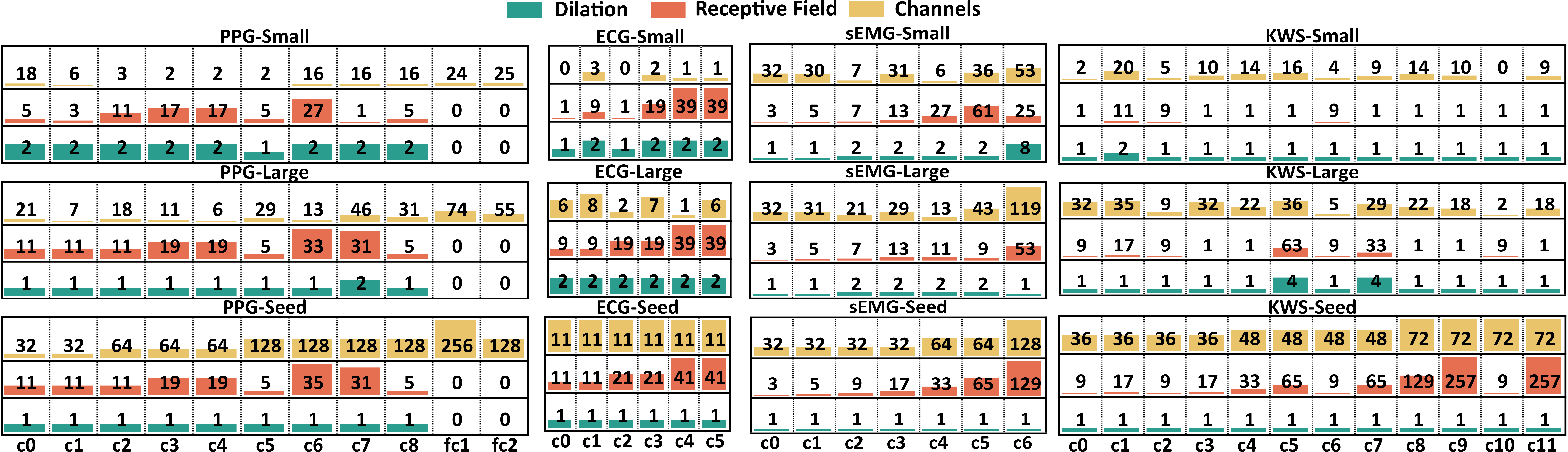}
  \caption{\rev{Hyperparameters of the deployed PIT architectures and corresponding seed network for the four benchmarks.}}\label{fig:found_arch}
\end{figure*}

Therefore, we select a coarser-grain search space for ProxylessNAS, trying to make the comparison with PIT as fair as possible, while keeping the search space size similar to the one of the original paper~\cite{proxylessnas_2018}.
To do so, we use the following procedure. \rev{First, we perform multiple ProxylessNAS searches on $C_{out}$, $F$ and $d$ separately, keeping the two not-optimized hyper-parameters at the seed values}. In each of these searches, we consider 4 layers variants in each super-net node, uniformly sampling the PIT search space \rev{(in the same way described above for FBNetV2-Coarse)}. \revRemove{For example, when searching for channels, we build variants of each layer with $\nicefrac{1}{4}C_{out,seed}$, $\nicefrac{1}{2}C_{out,seed}$, $\nicefrac{3}{4}C_{out,seed}$ and $C_{out,seed}$ output channels, while keeping receptive field and dilation to the seed values.} We then run ProxylessNAS multiple times with different regularization strengths. \revRemove{After repeating the same procedure for $F$ and $d$, we} \rev{We} identify, for every layer, the two values of each hyper-parameter that have been chosen more frequently. The $2^3$ possible combinations of the latter are used to generate the \textit{combined} search space for ProxylessNAS, which, accordingly, includes 8 layer variants in each super-net node.
The Pareto fronts of Figure~\ref{fig:proxyless_perf_comp} are obtained running ProxylessNAS multiple times on this combined search space, with different regularization strengths.
We report both the results obtained with the training scheme proposed in the original paper (\textit{Original} curve), which runs for a fixed number of epochs, and with the same early-stop mechanism employed for PIT (\textit{EarlyStop} curve). As shown, the quality of results is similar in both cases.

Figure~\ref{fig:timing_comparison} compares the search space dimension and average execution time of \revRemove{PIT, MorphNet and ProxylessNAS } \rev{the different tools} on the HR monitoring benchmark. For reference, the execution time of a standard training of the seed network is also reported. 
All time results refer to \rev{the search phase only (without warm up), and are obtained on} a single NVIDIA Titan XP GPU with a batch-size of 128.
For ProxylessNAS, we report the results of the the initial single-hyper-parameter searches (\textit{Proxyless-Single}) and of the final combined search (\textit{Proxyless-Multiple}), both with and without early-stopping.

Our algorithm explores a $10^{26}\times$/$10^{12}\times$ larger search space than Proxyless-Single/-Multiple. Further, it is only $1.13\times$ slower than Proxyless-Single with early-stopping, and $3.55\times$ faster than the variant without early-stopping, while it is $3.0\times$/$14.22\times$ faster compared to Proxyless-Multiple with/without the early-stopping training.
With respect to MorphNet, we explore a $10^{11}\times$ larger search space at the cost of a small $1.07\times$ increase in runtime.
\rev{FBNetV2-Coarse is the fastest tool, converging in few search epochs. Whereas offering a $2.5\times$ speedup with respect to PIT, the explored search space is $10^{26}$ smaller. Instead, FBNetV2-Fine explores a $10^{11}$ smaller space while requiring the same search time of the proposed approach}.
Lastly, \rev{PIT's} time-overhead \revRemove{of our approach} with respect to a normal training is only $34\%$.

\input{tables/01_table_deployment}

\subsection{Embedded Deployment} \label{subsec:deployment}
This section analyzes the results obtained deploying two TCNs for each target benchmark on the GAP8 IoT processor (running at \unit[100]{MHz}) and on the STM32H7 MCU (at \unit[480]{MHz}).
For each task, we deploy the best performing network in terms of MAE or accuracy (L). Moreover, we also select a small network that achieves a MAE drop $<1$ BPM, or an accuracy drop $<5\%$ with respect to the best performing one (S).
For comparison, we also deploy the baseline hand-tuned architectures (HT).
%
% All deployed TCNs are quantized to \texttt{int8} precision.
%
Table~\ref{tab:deployment} reports the results in terms of performance (MAE or accuracy, depending on the dataset), memory footprint, inference latency and energy consumption, while Figure~\ref{fig:found_arch} shows the hyper-parameters selected by our NAS.

PIT finds competitive solutions for both hardware targets and for all 4 tasks, despite the large difference in complexity among them, testified by the more than two orders of magnitude span in memory, latency and energy consumption in the results of Table~\ref{tab:deployment}.
For PPG-based HR monitoring, the L/S models achieve a 0/0.70 BPM MAE increase with respect to the hand-tuned TEMPONet respectively, while resulting in a 8.03/90.8$\times$ lower memory footprint, and a 5.45/19.6$\times$ lower latency and energy consumption on GAP8. On the STM32 MCU, the latency and energy reduction of the two PIT outputs is 3.83/18.2$\times$.
PIT's L/S models for ECG processing, instead, achieve +0.07\%/-1.36\% accuracy with respect to the hand-tuned ECGNET, with a 2.83/16.8$\times$ lower memory footprint, 2.13/3.44$\times$ latency and energy reduction on GAP8, and 
2.34/3.7$\times$ on the STM32.
For the sEMG gesture recognition task, the L/S models found by PIT obtain +2.31\%/-1.92\% accuracy compared to TCCNet. In this case, the higher accuracy of the large model is paid with a 3.57$\times$ larger memory footprint, and a 3.85$\times$ latency and energy increase on GAP8 (3.33$\times$ on the STM32H7), proving once again the goodness of the hand-tuned model for this task. The small TCN, instead, results in a 2.51$\times$ memory reduction, and 1.54$\times$ and 1.72$\times$ lower latency and energy on the two targets.
Lastly, the L/S PIT outputs for KWS obtain +0.16\%/-5\% accuracy with respect to TCResNet-14, with a 5.72/33.1$\times$ lower memory footprint, 3.58/9.54$\times$ lower energy and latency on GAP8, and 2.9/11.54$\times$ lower energy and latency on STM32H7.

Figure~\ref{fig:found_arch} shows the high variability of hyper-parameters settings found by PIT, and the different optimization behaviours for different benchmarks.
%. 
%
\revRemove{The two networks relative to the ECG benchmark are similar, but the small one eliminates the two residual layers by assigning them 0 channels.}
\rev{In general, comparing PIT outputs with the respective seeds, we can observe that the optimized networks found by our tool contradict several ``rules of thumb'' of manual DNN design, such as progressively increasing the number of channels and dilation for deeper layers. Accordingly, our NAS could also provide interesting insights for better TCN design. 
For instance, for the PPG benchmark, PIT finds solutions that are characterized by a large number of channels in the first and last layers, while keeping an overall high receptive field in the core of the network.}

%% file: tables/table_lambda.tex
\begin{table}[t]
\centering
\caption{Range of regularizer strength ($\lambda$) values for the four benchmarks.}\label{tab:range_lambda}
\begin{adjustbox}{max width=1\columnwidth}
\begin{tabular}{c|cccc}
Regularizer & PPG & ECG & sEMG & KWS \\ \cmidrule{1-5} 
$\mathcal{R}_{size}$ & $1\text{e-}7:5\text{e-}4$    & $5\text{e-}7:7.5\text{e-}3$    & $1\text{e-}7:5\text{e-}6$     & $5\text{e-}10:1\text{e-}5$    \\ \cmidrule{1-5}
$\mathcal{R}_{ops}$  & $1\text{e-}8:5\text{e-}5$    & $5\text{e-}8:5\text{e-}4$    & $5\text{e-}10:5\text{e-}8$     & $1\text{e-}10:1\text{e-}6$ \\\cmidrule{1-5}
\end{tabular}
\end{adjustbox}
\end{table}
%
% \begin{table}[h]
% \centering
% \caption{Range of regularizer strength ($\lambda$) values for the four benchmarks.}\label{tab:range_lambda}
% \begin{adjustbox}{max width=1\columnwidth}
% \begin{tabular}{ccccccccc}
% & \multicolumn{2}{c}{PPG} & \multicolumn{2}{c}{ECG} & \multicolumn{2}{c}{sEMG} & \multicolumn{2}{c}{KWS} \\ \cmidrule{2-9} 
% $\mathcal{R}_{size}$ & \multicolumn{2}{c}{$1\text{e-}7:5\text{e-}4$}    & \multicolumn{2}{c}{$5\text{e-}7:7.5\text{e-}3$}    & \multicolumn{2}{c}{$1\text{e-}7:5\text{e-}6$}     & \multicolumn{2}{c}{$5\text{e-}10:1\text{e-}5$}    \\ \cmidrule{2-9}
% $\mathcal{R}_{ops}$  & \multicolumn{2}{c}{$1\text{e-}8:5\text{e-}5$}    & \multicolumn{2}{c}{$5\text{e-}8:5\text{e-}4$}    & \multicolumn{2}{c}{$5\text{e-}10:5\text{e-}8$}     & \multicolumn{2}{c}{$1\text{e-}10:1\text{e-}6$}   
% \end{tabular}
% \end{adjustbox}
% \end{table}

%% file: tables/01_table_deployment.tex
\begin{table}[t]
\centering
\footnotesize
\caption{Detailed deployment results for the four benchmarks.}\label{tab:deployment}
\begin{adjustbox}{max width=1\columnwidth}
\begin{tabular}{|c|c|c|c|c|c|c|c|}
\cline{5-8}
\multicolumn{4}{c|}{} &  \multicolumn{2}{|c|}{GAP8} & \multicolumn{2}{|c|}{STM32} \\
\hline
 & & Perf. & Mem. & Lat. & En. & Lat. & En.\\
Task & TCN & int8 (float32) & [kB] & [ms] & [mJ] & [ms] & [mJ]\\\hline
%\multicolumn{1}{|l|}{TEMPONet seed} & 939K & 5.08 MAE & 112.6 ms & 29.5 mJ &  &  \\
\multirow{3}{*}{PPG} & HT & 5.01 (5.14) BPM & 423 & 23.2 & 1.2 & 58.3 & 13.6 \\
& S & 5.71 (6.17) BPM & 4.7 & 1.18 & 0.06 & 3.2 & 0.75\\
& L & 5.01 (5.03) BPM & 53.2 & 4.25 & 0.22 & 15.2 & 3.56\\ \hline \hline
%\multicolumn{1}{|l|}{ECGTCN seed} &  &  &  &  &  &  \\
\multirow{3}{*}{ECG} & HT & 94.2 (94.2) \% & 15.2 & 2.69 & 0.14 & 6.66 & 1.56\\
& S & 92.84 (93.16) \% & 0.9 & 0.78 & 0.04 & 1.8 & 0.42\\
& L & 94.13 (94.13) \% & 5.4 & 1.26 & 0.06 & 2.84 & 0.66\\ \hline \hline
%\multicolumn{1}{|l|}{TCCNet seed} &  &  &  &  &  &  \\
\multirow{3}{*}{sEMG} & HT & 88.89 (88.87) \% & 88.8 & 61.0 & 3.11 & 291 & 68.1\\
& S & 86.97 (86.98) \% & 35.4 & 39.6 & 2.02 & 169 & 39.5\\
& L & 91.2 (90.99) \% & 317.8 & 238 & 12.1 & 960 & 225\\ \hline \hline
%\multicolumn{1}{|l|}{TCResNet-14 seed} &  &  &  &  &  &  \\
\multirow{3}{*}{KWS} & HT & 92 (92.31) \% & 323.4 & 13.4 & 0.68 & 30.7 & 7.17\\
& S & 87 (86.58) \% & 9.8 & 1.40 & 0.07 & 2.66 & 0.62 \\
& L & 92.16 (92.64) \% & 56.5 & 3.74 & 0.19 & 10.6 & 2.48\\ \hline 
\end{tabular}
\end{adjustbox}
\end{table}

%% file: sections/06_Conclusions.tex
We have proposed PIT, a lightweight NAS tool for TCNs, able to explore a large, fine-grained search space of architectures with low GPU memory requirements. PIT is, to the best of our knowledge, the first DMaksingNAS tool explicitly designed for 1D convolutional networks, and the first to target the optimization of the receptive field and dilation of convolutional layers.
With experiments of four real-world benchmarks, we have shown that PIT is able to find improved  versions  of  state-of-the-art TCNs, with a memory compression of up to 8.03$\times$ (90.8$\times$) and a latency and energy reduction of up to 5.45$\times$ (19.6$\times$) without (with a reasonable) accuracy drop, when deployed on commercial edge devices.
\rev{Our future work will focus on extending PIT principles to generic N-dimensional CNNs.}